\documentclass[]{interact}

\usepackage{epstopdf}
\usepackage[caption=false]{subfig}
\usepackage[nolists,tablesfirst]{endfloat}

\usepackage[numbers,sort&compress]{natbib}
\bibpunct[, ]{[}{]}{,}{n}{,}{,}
\renewcommand\bibfont{\fontsize{10}{12}\selectfont}

\theoremstyle{plain}

\theoremstyle{definition}

\theoremstyle{remark}

\usepackage{amsmath}
\usepackage{natbib}

\usepackage{url}
\usepackage{bm}
\usepackage{amsfonts} 
\usepackage{commath} 
\usepackage{tabularx, booktabs}
\usepackage{makecell}
\usepackage{adjustbox}
\usepackage{graphicx}
\usepackage{placeins}
\usepackage{hyperref}
\usepackage{cleveref}
\usepackage{appendix}
\usepackage{amsmath,epsfig,epsf,psfrag, url, setspace, graphicx,mathtools,comment}
\usepackage{amssymb, multirow}
\usepackage{xcolor,colortbl}
\usepackage{arydshln} 
\usepackage{dsfont} 

\usepackage{etoolbox}
\makeatletter
\patchcmd{\@makecaption}
  {\parbox}
  {\advance\@tempdima-\fontdimen2} 
  {}{}
\makeatother  
\usepackage{enumerate}
\usepackage{float}
\newcommand{\defeq}{\vcentcolon=}

\begin{document}

\articletype{Original Research Article}

\title{Multi-resolution super learner for voxel-wise classification of prostate cancer using multi-parametric MRI}

\author{
\name{Jin Jin\textsuperscript{a}\thanks{CONTACT Jin Jin. Email: jjin31@jhu.edu. Department of Biostatistics, Bloomberg School of Public Health, Johns Hopkins University, Baltimore, MD 21205, USA}, 
Lin Zhang\textsuperscript{b},
Ethan Leng\textsuperscript{c},
Gregory J. Metzger\textsuperscript{d},
Joseph S. Koopmeiners\textsuperscript{b}}
\affil{
\textsuperscript{a}Department of Biostatistics, Bloomberg School of Public Health, Johns Hopkins University, Baltimore, MD, USA; 
\textsuperscript{b}Devision of Biostatistics, School of Public Health, University of Minnesota, Minneapolis, MN, USA;
\textsuperscript{c}Department of Biomedical Engineering, University of Minnesota, Minneapolis, MN, USA;
\textsuperscript{d}Department of Radiology, University of Minnesota, Minneapolis, MN, USA.
}
}

\maketitle

\begin{abstract}
Multi-parametric MRI (mpMRI) is a critical tool in prostate cancer (PCa) diagnosis and management. To further advance the use of mpMRI in patient care, computer aided diagnostic methods are under continuous development for supporting/supplanting standard radiological interpretation. While voxel-wise PCa classification models are the gold standard, few of any approaches have incorporated the inherent structure of the mpMRI data, such as spatial heterogeneity and between-voxel correlation, into PCa classification. We propose a machine learning-based method to fill in this gap. Our method uses an ensemble learning approach to capture regional heterogeneity in the data, where classifiers are developed at multiple resolutions and combined using the super learner algorithm, and further account for between-voxel correlation through a Gaussian kernel smoother. It allows any type of classifier to be the base learner and can be extended to further classify PCa sub-categories. We introduce the algorithms for binary PCa classification, as well as for classifying the ordinal clinical significance of PCa for which a weighted likelihood approach is implemented to improve the detection of less prevalent cancer categories. The proposed method has shown important advantages over conventional modeling and machine learning approaches in simulations and application to our motivating patient data.
\end{abstract}


\section{Introduction}\label{ch1}
Prostate cancer (PCa) is the second most common cancer and the second leading cause of cancer death among men in the U.S. 
\cite{WinNT}. 
In recent years, Multi-parametric magnetic resonance imaging (mpMRI) has become an increasingly important tool for evaluating the extent of PCa and determining the corresponding management strategy, such as needle biopsy, prostatectomy, and focal therapy \cite{Dickinson2011,hegde2013}. 
Traditionally, mpMRI examinations are used to manually delineate the cancerous regions within the prostate, 
which requires high efficiency due to time constraints and depends highly on radiologists' and urologists' expertise. 
Such methods have been criticized for the large variability in radiological assessment and human error \cite{Garcia-Reyes2015}, which motivated the development of automatic, quantitative predictive methods that address the limitations of direct radiological interpretation \cite{Lemaitre2015}. Current methods for voxel-wise PCa classification mainly focus on binary classification, i.e. to distinguish cancerous prostate voxels from the benign ones. Representative methods include regression-based models, 
clustering methods, kernel methods, naive Bayes, neural network, graphical models, and other commonly implemented machine learning-based approaches \cite{puech2009,artan2010,Mazzetti2011,Niaf2012,Shah2012,Peng2013,Matulewicz2014,Litjens2014,fei2017computer,trigui2017automatic,zhu2017mri,algohary2018radiomic,viswanath2019comparing,zhu2019clustering,de2020deep,gholizadeh2020voxel,mazzetti2018computer,zhu2020high,zhu2021time,jarrow2021low,mehta2021computer}.
Textural feature models have also been developed \cite{add3,Khalvati2015,toivonen2019radiomics}. 

Previous research has revealed important regional heterogeneity of the prostate mpMRI data where there are substantial differences in the distribution of both the predictors (the observed mpMRI parameters such as the apparent diffusion coefficient; ADC) and the outcome (voxel-wise cancer status), across two main zones of the prostate: the central gland (CG) and the peripheral zone (PZ), and additional local features that lead to complex heterogeneous data patterns across the whole gland. There is also strong between-voxel correlation observed within each prostate. Conventional machine learning algorithms, such as random forest and support vector machine\cite{artan2010,parfait2012,trigui2017automatic,zhu2017mri}, tend to treat mpMRI data of all prostate voxels as independent and identically distributed observations and are unable to specifically model these complex mpMRI data features, which may have negative impact on the classification results. Although there have been attempts to account for heterogeneity across the two main zones of a prostate \cite{trigui2017automatic,jin2018,viswanath2019comparing,gholizadeh2020voxel}, there is little discussion on developing a model that can simultaneously account for the various, complex structures of mpMRI data. Given the typically small sample sizes for such mpMRI studies\cite{Lemaitre2015,stabile2020multiparametric}, machine learning approaches with tremendous flexibility, such as the black-box algorithms, are likely to encounter over-fitting issues without modeling the specific data structures that can be critical for PCa detection.

To address such limitations of the conventional classifiers, we recently developed a Bayesian hierarchical modeling framework that can take into account heterogeneity in the distribution of mpMRI data between the PZ and CG for improved voxel-wise classification of binary PCa status \cite{jin2018}. 
Following a similar Bayesian modeling framework, scalable modeling of the between-voxel correlation in the mpMRI data has also been discussed \cite{jin2020}. Although having provided promising tools for modeling the complex structures of high-dimensional mpMRI data, these Bayesian methods have also shown limitations. For example, despite having implemented dimension reduction techniques, the modeling of the between-voxel correlation is still computationally intensive. It is thus not ideal to extend these models to further incorporate more complex local features that cause spatial heterogeneity across the whole prostate. Additionally, there is a need to extend binary PCa classification to more complex classification problems, such as detailed categorization of the clinical aggressiveness of PCa based on Gleason score \cite{epstein2005,delahunt2012}, which will be computationally challenging for Bayesian hierarchical models due to the large number of additional model parameters involved.

In this paper, we propose a machine learning-based approach to voxel-wise classification of PCa, which flexibly incorporates the various mpMRI data structures and addresses limitations of the recently proposed Bayesian classification models. 
The main feature of the proposed method is a multi-resolution modeling approach using ensemble learning via the super learner algorithm \cite{Van2007}. This approach enables our method to capture both homogeneous/global features and heterogeneous/local features of mpMRI, which is a fundamental improvement compared to previous Bayesian classifiers that can only handle between-anatomic-zone heterogeneity assuming a stationary spatial process. 
Briefly, we first train a given classification model/algorithm (which we call ``base learner'') locally within each sub-region of the prostate under different resolutions. The multi-resolution, sub-region specific learners are then combined for voxel-wise classification of future data. 
The super learner framework also allows the implementation of any type of base learners, which means that the resulting classifier can be continuously updated when novel, superior classification methods are available. 
We propose to account for between-voxel correlation by applying a spatial Gaussian kernel smoother to the voxel-wise cancer probabilities predicted by the multi-resolution base learners \cite{nadaraya1964,watson1964,jin2018}. As opposed to the previous Bayesian spatial modeling framework, this spatial smoothing technique is considered due to practical concerns including its minimal computational cost, ease of implementation, and generally good performance in reducing random noise in the data. 
We will first introduce the algorithm for classifying binary PCa status. To demonstrate the flexibility of the method in its extension to classifying more complex PCa outcomes, we then discuss how to modify the method to classify the clinical significance of PCa based on Gleason score, which is an ordinal outcome that measures the aggressiveness of PCa. 
Simulation studies and application to our motivating data set were conducted for both binary and ordinal classification, which illustrate the advantages of our proposed method relative to several commonly implemented machine learning approaches, indicating the potential of the resulting classifiers to provide enhanced guidance for clinical decision making.

The rest of the paper is organized as follows. In Section \ref{sec2}, we briefly introduce our motivating data set and notations. In Section \ref{sec3}, we introduce our proposed method and the corresponding algorithm for the binary classification of PCa status, the performance of which is illustrated in Section \ref{sec4} through simulation studies and application to patient data. 
In Section \ref{sec5}, we propose the algorithm for classification of the ordinal, clinical significance of PCa, and discuss its performance on the synthetic and the real patient data. Section \ref{sec6} summarizes the paper and provides a discussion for future directions.

\section{Voxel-wise MpMRI data and notations}
\label{sec2}
We first give an overview of our motivating data, which were collected on the voxel level from 34 prostate slices of 34 different patients diagnosed with PCa \cite{Metzger2016}. Briefly, maps of voxel-wise quantitative MRI parameters, including apparent diffusion coefficient (ADC), area under the gadolinium concentration time curve at 90 seconds (AUGC90), reflux rate constant ($\text{k}_{\text{ep}}$), forward volume transfer constant ($\text{K}^{\text{trans}}$), fractional extravascular extracellular space ($\text{V}_{\text{e}}$), 
and T2 values, 
were calculated. 
Quantitative T2 maps were calculated from fast spin echo data sets acquired at additional echo times\cite{Liney1996}.
Diffusion-weighted imaging (DWI) and dynamic contrast enhanced (DCE) MRI, were acquired as part of a clinical multiparametric magnetic resonance imaging (mpMRI) exam. ADC maps were calculated from DWI data acquired with 3 diffusion-encoding b-values of 50, 400 and 1000. Pharmacokinetic maps were generated from DCE-MRI data using a modified Tofts model with a customized population-averaged arterial input function \cite{Metzger2016}; maps of the forward volume transfer constant ($\text{K}^{\text{trans}}$), reflux rate constant ($\text{k}_{\text{ep}}$), and area under the gadolinium concentration time curve at 90 seconds (AUGC90) were calculated.

\begin{figure}[ht!]
\centering
\includegraphics[width=270pt,height=16.2pc]{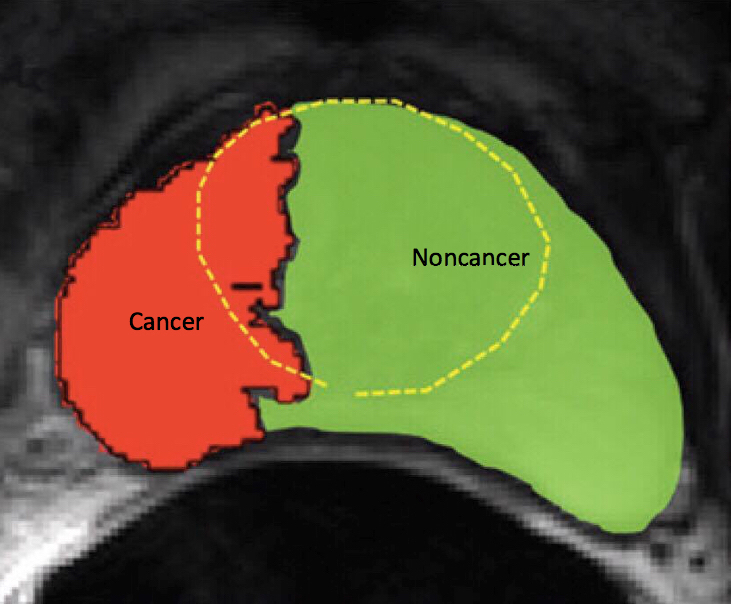}
\caption{Annotated voxel-wise cancer status of an example prostate slice. Red and green indicate annotated cancer and noncancer voxels, respectively. The yellow dashed curve divides the prostate gland into peripheral zone (PZ, the area inside the curve) and central gland (CG, the area outside the curve).
\label{fig2.1}
}
\end{figure}

The mpMRI parameters used in our classification problem include ADC, AUGC90, $\text{K}^{\text{trans}}$ and $\text{k}_{\text{ep}}$: this is the combination that gives the best PCa classification results based on the least absolute shrinkage and selection operator (lasso) model in \cite{Metzger2016}. 
Note that the ADC values are approximately normally distributed, while the right-skewed AUGC90, $\text{K}^{\text{trans}}$ and $\text{k}_{\text{ep}}$ values are log transformed to have an approximately normal distribution. 
For the data used in this series, all datasets were manually co-registered to support the development of the proposed voxel-wise model.
Manually guided annotation for the zonal information (the location of PZ and CG) was conducted on the T2-weighted images. After surgery, the pathologists manually annotated the cancerous areas and their Gleason scores on the histopathology slides, the maps of which were then co-registered with the corresponding maps of the various MRI parameters \cite{kalavagunta2015registration}. Figure \ref{fig2.1} shows the image of a prostate slice annotated with voxel-wise binary cancer status and location of the two main zones (PZ and CG). 
\FloatBarrier

Current research on voxel-wise detection of PCa using mpMRI mainly focus on binary classification where prostate voxels are classified as benign or cancerous. 
However, more refined classification of cancerous voxels, such as distinguishing between sub-categories of PCa according to their level of clinical significance, can provide enhanced guidance for clinical applications. 
A widely used system for determining PCa aggressiveness is the Gleason grading system, which was originally proposed in 1966, refined in 1977 gaining almost universal acceptance, then updated in 2005 to a version that has been widely used since then \cite{epstein2005,delahunt2012}. The Gleason grade, which ranges from 3 to 5, describes the degree of abnormality for the organization of PCa cells on histologic examination (5 indicates the most abnormal). 
Each patient is assigned with two Gleason grades: a primary score, $S^a$, that describes the predominant pattern, and a secondary score, $S^b$, that describes the second predominant pattern \cite{jensen2019}. The total Gleason score, $S^a + S^b$, 
is then used to evaluate PCa aggressiveness of the patient. 
Gleason score 7 tumors show heterogeneity in the biological behavior, with clear differences in prognosis between patients with score 3+4 tumors and patients with score 4+3 tumors at radical prostatectomy \cite{chan2000}. We then utilize this to categorize cancerous tissues into clinically insignificant cancer ($S^a + S^b = 3+3$ or $3+4$) and clinically significant cancer ($S^a \geq 4$). This results in three ordinal categories of PCa: no PCa, clinically insignificant PCa and clinically significant PCa, which is consistent with newly proposed guidelines for deciding men's eligible for active surveillance \cite{epstein2010,matoso2019}. 

We now introduce notations which will be used to describe the voxel-wise mpMRI data. Assume that there are $N=34$ patients (i.e. prostate slices), and $n_i$ ($2098\leq n_i \leq 5756$) voxels in the image of the $i$-th slice, $i=1,\ldots,N$. For the $j$-th voxel in the $i$-th image, a $d\times1$ vector of quantitative mpMRI parameters, $\bm{y}_{ij}=(y_{ij,1},...,y_{ij,d})^T$, is measured. 
For our motivating data set, we only consider $d=4$ mpMRI parameters including ADC, AUGC90, $\text{k}_{\text{ep}}$ and $\text{K}^{\text{trans}}$, the combination of which provides the highest average area under the ROC curve (AUC) using the generalized linear model in \cite{Metzger2016}. 
Each voxel is annotated with a primary and a secondary Gleason score, which are denoted as $S^a_{ij}$ and $S^b_{ij}$, respectively, based on which we define a binary cancer indicator, $c_{ij}=\mathds{I}(S^a_{ij}+S^b_{ij}\geq 6)$ (1: cancer, 0: noncancer), where $\mathds{I}(\cdot)$ is the indicator function, and an ordinal outcome for the clinical significance of PCa, $G_{ij}$, which equals 1 (noncancer) if $S^a_{ij}\leq 3$ and $S^a_{ij}+S^b_{ij}<6$, 2 (clinically insignificant cancer) if $S^a_{ij}+S^b_{ij} \in \{3+3, 3+4\}$, and 3 (clinically significant cancer) otherwise. The location information of the voxel is described by $r_{ij}$, an indicator for zone (1: PZ, 0: CG), and $\bm{s}_{ij}$, a 2-D coordinate standardized across all prostate images. Note that there is currently no standard template for prostate due to the variability in size, shape, etc., and therefore we conducted a rough rescaling on the original voxel-wise coordinates to ensure that all images fall into the same support $(-1,1)\times(-1,1)$, with $(0,0)$ being the center of each prostate slice.

The voxel-wise distribution of the various mpMRI parameters for the voxels of each Gleason score, each binary PCa status, and each category of the clinical significance of PCa, are summarized in Supplementary Figures S1 - S3. 
The sample prevalences of the various Gleason grade groups are: $p_{3+3}=0.23$, $p_{3+4}=0.035$, $p_{3+5}=0.005$, $p_{4+3}=0.020$, $p_{4+4}=0.022$, $p_{4+5}=0.044$,  $p_{5+4}=0.017$, and  $p_{5+5}=1.9\times 10^{-4}$. The total sample prevalence of PCa voxels is 0.167; the sample prevalences of the clinically insignificant PCa voxels (``3+3'' or ``3+4'') and clinically significant PCa voxels (``3+5'', ``4+3'', ``4+4'', ``4+5'', ``5+4'', or ``5+5'') are 0.058 and 0.109, respectively. 
Note that these summaries are not fully informative, because they only provide information on the voxel-wise distribution of the mpMRI data but not the regional heterogeneity or local spatial patterns of the data that we aim to capture.

\section{Methods}
\label{sec3}
Our method was motivated by the need for a flexible and computationally efficient PCa classification approach that can incorporate complex structures of the mpMRI data. Specifically, we propose a two-stage model that incorporates the heterogeneity in the distribution of mpMRI data across the prostate gland via a super learner framework \cite{Van2007}. In stage one, we first select a base learner, for example, a statistical model or a machine learning algorithm that can classify the voxel-wise PCa status. Next, we segment $(-1,1)\times(-1,1)$, the 2-D support of the prostate gland, 
into $k\times k$ equal-size, rectangular sub-regions using a set of different values for $k\in \mathbb{N}^{+}$. Under the $k\times k$ resolution, we train the selected base learner locally in each of the $k^2$ sub-regions.  
In stage two, we use the classification results from the stage-one, multi-resolution base learners to train a new classifier, which is essentially a weighted combination of the base learners.

\subsection{Standard super learner algorithm}\label{sec3.1}
We first provide a brief overview of the standard super learner algorithm, which was originally proposed as an ensemble learning-approach for prediction \cite{Van2007}. Super learner constructs an optimal weighted combination of multiple candidate learners using Cross-Validation, which has been shown to perform asymptotically as well as the oracle learner (i.e. the learner that minimizes risk under the true data-generating distribution) in terms of expected risk difference among the family of candidate learners, if the number of candidate learners, $K$, is polynomial in sample size, $n$, i.e. $K\leqslant n^q$ for some $q<\infty$ \cite{Van2007}. 
Intuitively, the super learner uses ensemble learning to ``average'' across multiple candidate learners, therefore capturing the characteristics of the data that are revealed by various types of prediction methods.

The standard super learner algorithm proceeds as follows. 
Suppose we have $n$ i.i.d. observations, $(X_i, Y_i)\sim F_0$, $i = 1,2,\ldots, n$. The goal is to train a regression model, $\widehat{\Psi}_0(X) = E_0(Y|X)$, 
which is the minimizer of the expectation of a loss function, $E_0 L(X,Y,\Psi)$. Note that the super learner is applicable to any parameter (prediction model) that can be defined as the minimizer of a loss function over a parameter space. 
Assume that there are $K$ candidate prediction models, $\Psi_k$, $k=1,2,\ldots,K$, each representing a different mapping from the data, $P_n =\{(Y_i,X_i),i=1,2,\ldots,n\}$, to the parameter space of the functions of $X$. The super learner utilizes a two-stage modeling framework. First, the selected prediction models are trained on the entire data set to obtain the stage-one prediction models, $\{\widehat{\Psi}_{k} \defeq \widehat{\Psi}_k(P_{n}),k=1,2,\ldots,K\}$. Next, a stage-two model is fitted to construct a weighted combination of the prediction results of the stage-one models. Specifically, a V-fold Cross-Validation is conducted to determine the weight of each candidate learner: suppose that $v\in \{1,2,\ldots,V\}$ denotes a split of the sample that generates an index set for the training sample, $T(v)$, and an index set for the validation sample, $V(v)$, where $T(v) \bigcup V(v) = \{1,2,\ldots,n\}$, $\bigcup_{v=1}^V V(v) = \{1,2,\ldots,n\}$ and $V(v_1)\bigcap V(v_2) = \emptyset$, $\forall v_1 \neq v_2$.
For each $v$, we obtain $\widehat{\Psi}_{k}^v \defeq \widehat{\Psi}_k(P_{n,T(v)})$, i.e., the realization of each $\Psi_k$ on the training set $P_{n,T(v)}$, then apply $\{\widehat{\Psi}_{k}^v,k=1,2,\ldots,K\}$ to the corresponding validation set $V(v)$ to obtain predictions $\{Z_{i} = (\widehat{\Psi}_{1}^v (X_i), \ldots, \widehat{\Psi}_{K}^v (X_i))^T, v: i\in V(v)\}$, so that each observation $i$ will have a vector of $K$ Cross-Validated predictions, $Z_{i}$, obtained from the $K$ candidate models trained based on the $v$-th split in Cross-Validation where the $i$-th sample is in the validation set. A stage-two model, $\hat{\Phi}: \mathcal{Z}\rightarrow \mathcal{Y}$, is then trained based on  $\{(Y_i,Z_i),i=1,2,\ldots,n\}$, with $Z_i$ being the new covariates. 
The super learner then combines the trained stage-one models, $\{\widehat{\Psi}_{k} \defeq \widehat{\Psi}_k(P_{n}),k=1,2,\ldots,K\}$, by the stage-two model, $\hat{\Phi}$, to construct the final prediction model.

\subsection{The proposed algorithm for binary classification of PCa}\label{sec3.2}
Our proposed voxel-wise classification algorithm for PCa adopts the benefit of super learner to account for regional heterogeneity in the mpMRI data.
The basic framework is similar to that of the original super learner, but, instead of combining various types of learners, we combine learners of the same type but trained at different resolutions. 

\begin{figure}
\centering
   \includegraphics[scale=0.5]{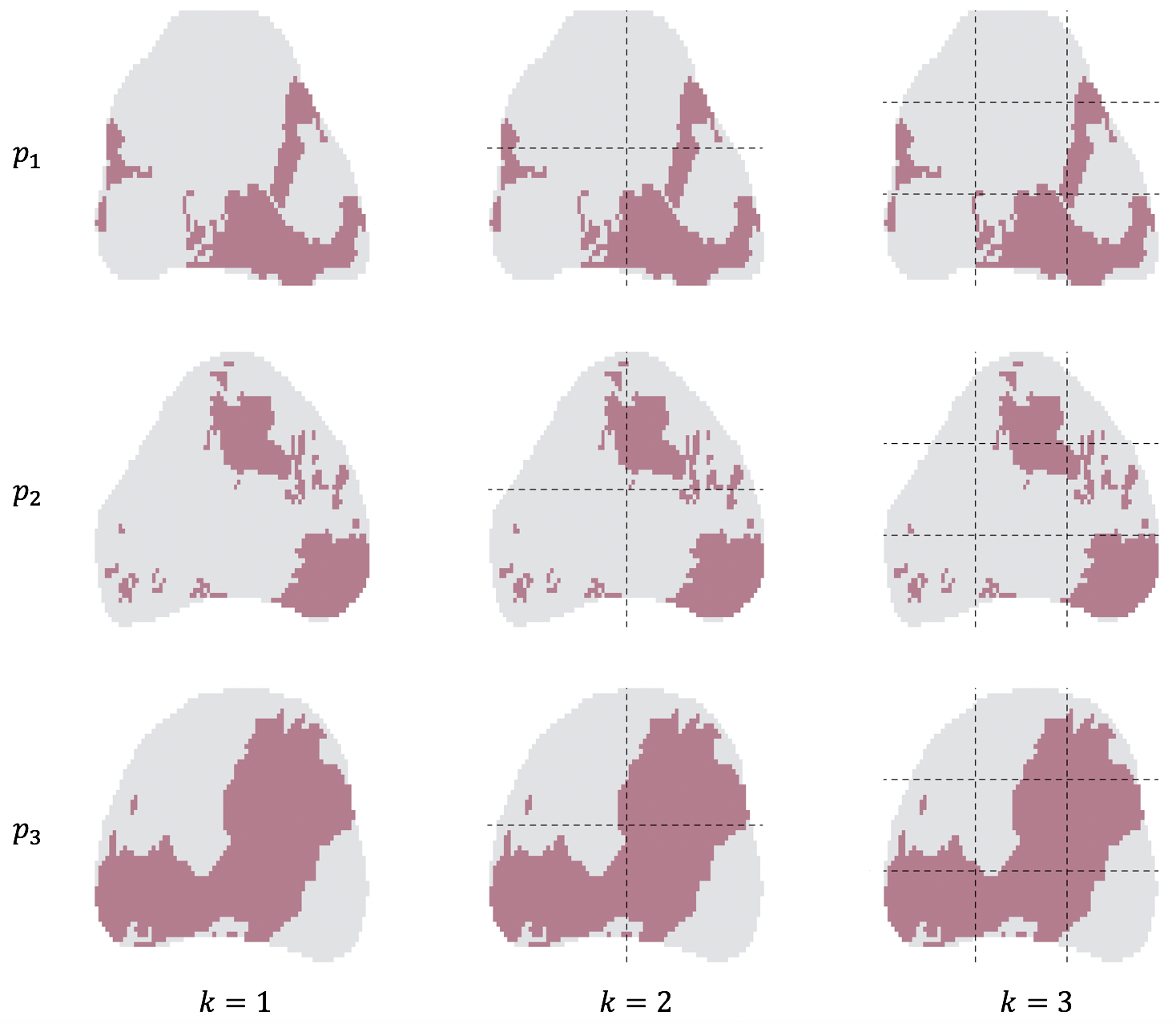}
\caption{Region segmentation for three example prostates ($p_1$, $p_2$ and $p_3$) under resolutions $k\times k$, $k\in\{1,2,3\}$. 
}
\label{fig.segmentation}
\end{figure}

Suppose the outcome of interest is the voxel-wise, binary PCa status, $c_{ij}$s. 
We first select a base learner, $\Psi$, which can be any model/algorithm applicable to the classification problem. 
We split the $N$ subjects into $V$ folds, and for the $v$-th split, $v=1,2,\ldots,V$, we divide the subjects into a training set with index $T(v)\subset \{1,2,\ldots,N\}$, that includes data in all but the $v$-th fold, and a validation set with index $V(v)=\{1,2,\ldots,N\}\backslash T(v)$. 
Under the $v$-th split, we first train the base learner using training data on the whole prostate gland (``WG''), which we denote as $A_{1,1}$, and obtain the global learner, $\widehat{\Psi}^v_{1,1}(\bm{y})$, where $\bm{y}$ denotes the vector of the mpMRI parameters. Note that the $\bm{y}$ here denotes the vector of covariates used in our classification problem, which is not related to the notation $Y$ in Section \ref{sec3.1}. 
Second, we segment the 2-D support of WG, $(-1,1)\times (-1,1)$, into $2\times 2$ equal-size sub-regions: $A_{2,1}=(-1,0)\times (-1,0)$, $A_{2,2}=(-1,0)\times (0,1)$, $A_{2,3}=(0,1)\times (-1,0)$, and $A_{2,4}=(0,1)\times (0,1)$. The voxels that fall onto the edge of some sub-regions can be included in any sub-region that share the edge.
We train the base learner separately within each sub-region to obtain region-specific learners, $\{\widehat{\Psi}_{2,l}^v,l=1,2,\ldots,2^2\}$. The trained learner under $2 \times 2$ resolution then becomes  $\widehat{\Psi}_2^v(\bm{y})=\sum_{l=1}^4 \mathds{I}(\bm{s}\in A_{2,l})\widehat{\Psi}_{2,l}^v(\bm{y})$, where $\bm{s}$ denotes the coordinate of a voxel. 
Third, we segment the support of WG into $3\times 3$ equal-size sub-regions: $\{A_{3,l}=(a,a+2/3)\times (b,b+2/3),a,b = -1,-1/3,1/3,l=1,2,\ldots,3^2\}$, then train the base learner within each region to get $\{\widehat{\Psi}_{3,l}^v,l=1,2,\ldots,3^2\}$. The trained learner under $3\times 3$ resolution is then $\widehat{\Psi}_3^v(\bm{y})=\sum_{l=1}^9 \mathds{I}(\bm{s}\in A_{3,l})\widehat{\Psi}_{3,l}^v(\bm{y})$. Theoretically, we can continue this segmentation and model training process with $k=4,5,6$, etc. 
In this paper, we consider $k \in \{1,2,3\}$, i.e., the lowest $K=3$ resolutions, for illustration. Figure \ref{fig.segmentation} illustrates the region segmentation on three example prostate images in our motivating dataset.
\FloatBarrier

After training the model under $k \times k$ resolution, we apply $\{\widehat{\Psi}_k^v,v=1,2,\ldots,V\}$ to their corresponding validation sets, which gives the Cross-Validated classification results,  
$x^k_{ij} \defeq \sum_{v=1}^V \mathds{I}(i\in V(v)) \widehat{\Psi}^v_k(\bm{y}_{ij})$. 
Before combing results at different resolutions, we would like to take into account the spatial dependency between voxels. 
Formal spatial modeling can be conducted, which, however, has a substantial computational burden even with appropriate dimension reduction techniques \cite{jin2020}. 
To avoid such computational cost, we propose to implement a computationally efficient spatial smoothing technique instead. Specifically, we apply the Nadaraya-Watson estimator with Gaussian kernel \cite{nadaraya1964,watson1964} 
to $\{x^k_{ij},j=1,2,\ldots,n_i\}$ separately for each image $i$, and obtain a vector of spatially smoothed, Cross-Validated voxel-wise classification results for each voxel: $\bm{\widetilde{x}}_{ij} = (\widetilde{x}^1_{ij},\widetilde{x}^2_{ij},\ldots,\widetilde{x}^K_{ij})^T$.
We then develop a stage-two classification model, $\Phi$, using $\bm{\widetilde{x}}_{ij}$s as the inputs.
Given that the outcome is binary, we consider a generalized linear model with a probit link function. 
Finally,  
we train the stage-one, multi-resolution base learners, $\Psi_k$s, on the entire data set, which, combined with the intermediate spatial smoothing step and the trained stage-two model $\widehat{\Phi}$, constructs the final classifier. 
\begin{figure}
\centering
   \includegraphics[width=\textwidth]{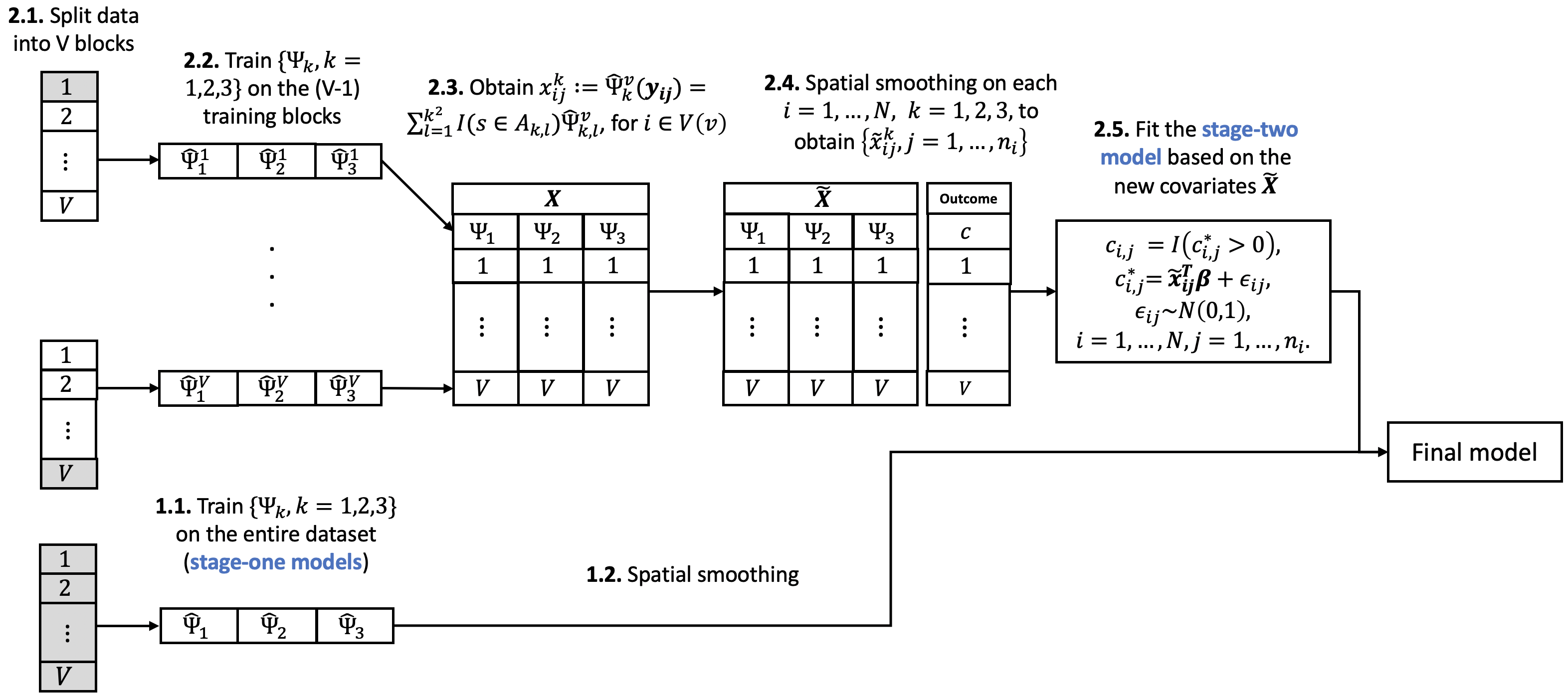}
	\caption{Workflow of the proposed classification algorithm for voxel-wise binary PCa status, $c_{ij}$s. ${\Psi}_{k}$ denotes the base learner under resolution $k$, 
    $\hat{\Psi}_{k,l}^v$ denotes the learner trained in the $l$-th sub-region, $A_{k,l}$, under resolution $k$ based on the $v$-th split of Cross-Validation, and $\hat{\Psi}_{k}^v$ summarizes the results from $\hat{\Psi}_{k,l}^v$, $l=1,2,\ldots,k^2$. The total number of resolutions used for prostate segmentation is set to $K=3$ for illustration.
	}
\label{sl-algorithm}
\end{figure}
\FloatBarrier

Figure \ref{sl-algorithm} summarizes the workflow of the proposed binary classification algorithm. We can observe two attractive features of the algorithm. 
First, it efficiently accounts for both global and complex local mpMRI structures via multi-resolution modeling. 
Second, it can implement any classification methods, even ``black-box'' machine learning algorithms, as the base learner. 
Currently, we only consider combining the three lowest resolutions for illustration, but the segmentation process can continue to finer resolutions depending on the original resolution (the number of voxels) of the mpMRI images and the locality of region heterogeneity. 
A practical concern is that, under finer-resolution segmentations, some sub-regions may have voxels of only one cancer status (cancer or non-cancer), making the model training challenging, and thus the upper limit of the resolutions for region segmentation should be chosen to minimize the number of such single-class sub-regions. 
However, we would like to note that with only a few single-class sub-regions, the algorithm can still be implemented: suppose that under segmentation of the highest resolution $K\times K$, all voxels in one sub-region $A_{K,l^{\ast}}$ are of the same cancer status $c\in \{0,1\}$. We then define $\widehat{\Psi}_{K,l^{\ast}}(\bm{y})\equiv c$, i.e., new voxels in this sub-region will be classified as $c$ with probability 1. Note that this classification result will be averaged with the results from lower-resolution classifiers that do not have this single-class issue, resulting in a classifier that is not uniformly equal to $c$ for all voxels in this sub-region.

\section{Application\label{sec4}}
\subsection{Simulation studies} \label{sec4.1}
We first conducted simulations to illustrate the performance of the proposed binary classification algorithm.  
We considered multiple choices for the base learner, including generalized linear model with probit link function (``GLM''), quadratic discriminant analysis (``QDA'', \cite{tharwat2016}) and random forest (``RF'', \cite{hastie2009}). Several other popular machine learning methods, such as the support-vector machine (SVM), were not included due to the computational burden, which made them infeasible with our data. Based on each base learner, we then applied the following classifiers: (1) ``Baseline'': the base learner; (2) ``SL0'': the proposed algorithm without the spatial smoothing step; and (3) ``SL'': the proposed algorithm. We assessed the improvement due to the proposed multi-resolution modeling strategy by comparing between classifiers (1) and (2), and the improvement due to spatial smoothing by comparing between classifiers (2) and (3). 
Additionally, we considered another classifier, ``GLM + QDA + RF'', which combined results from the multi-resolution GLM, QDA and RF all together using the proposed algorithm. For this classifier, we 
also considered either implementing or not implementing spatial smoothing (``SL'' and ``SL0'', respectively). 

The synthetic data were generated following the simulation procedure in \cite{jin2020}, with an additional step of adding random shifts in sub-regions of different resolutions. 
To reduce computational burden for the simulation study, we generated reduced-size images by taking every third row and column of the full-size images in our data, which resulted in about 300 to 800 voxels per image.
The shapes of the simulated prostate images, including the voxel-wise zone indicators $r_{ij}$s and standardized 2-D coordinates $\bm{s}_{ij}$s, were selected with replacement from those of the images in the motivating data set. Within each image, the voxel-wise cancer status and mpMRI parameters were simulated according to model (\ref{simu-model-paper3}):
\begin{align}
\bm{w}_i \sim 
\mathcal{MVN}(\bm{0},\bm{C}(\bm{S}_i,\bm{S}_i| & \bm{\theta})), \text{  }
c^{\ast}_{ij} \sim N(q_{r_{ij},0} + w_{ij},1),\nonumber\\
c_{ij} = I (c^{\ast}_{ij}>0),\text{  } &
\bm{e}^k_{i_k} \sim \mathcal{MVN} (\bm{0},\bm{\Lambda}), \nonumber\\
\bm{y}_{ij} \stackrel{ind}{\sim} \mathcal{MVN} ( \bm{\mu}_{c_{ij},r_{ij}} & +   \sum_{k=1}^K\bm{e}^k_{i^k_{ij}} + \bm{\delta}_i, \bm{\Gamma}_{c_{ij},r_{ij}}).
\label{simu-model-paper3}
\end{align}
Examples of the simulated mpMRI and PCa maps are summarized in Supplementary Figure S4. Briefly, to introduce spatial correlation between voxel-wise cancer status within each image $i$, we simulated a vector of spatially correlated random effects, $\bm{w_i}=(w_{i,1},\ldots,w_{i,n_i})^T$, from a multivariate normal distribution assuming a Mat\'ern correlation structure, i.e., the $(i,j)$-th entry of the spatial covariance matrix, $\bm{C}(\bm{S}_i,\bm{S}_i|\bm{\theta})$, was defined as 
$\bm{C}(\bm{s}_{ij},\bm{s}_{ik}|\bm{\theta}) =  \frac{\sigma^2}{2^{\nu-1} \Gamma({\nu})} \Big(\frac{2\nu^{1/2}\norm{\bm{s}_{ij}-\bm{s}_{ik}}}{\phi}\Big)^{\nu}
\times \bm{J}_{\nu} \Big(\frac{2\nu^{1/2}\norm{\bm{s}_{ij}-\bm{s}_{ik}}}{\phi}\Big)$, where $\bm{\theta} = \{\sigma^2,\phi,\nu\}$, and $\sigma^2$, $\phi$ and $\nu$ are the parameters that control the variance, range and smoothness of the spatial covariance among voxels, respectively. 
Second, we simulated $c_{ij}^{\ast}$s independently from $N(q_{r_{ij},0} + w_{ij})$s, where $q_{r_{ij},0}$ denotes the probit of cancer prevalence in zone $r_{ij}\in \{0,1\}$, and then simulated $c_{ij}=\mathds{I}(c^{\ast}_{ij}>0)$ according to the probit model. We assumed that the distribution of the mpMRI parameters, $\bm{y}_{ij}$, varied by cancer status $c_{ij}$ and zone indicator $r_{ij}$ with mean $\bm{\mu}_{c_{ij},r_{ij}}$ and covariance $\bm{\Gamma}_{c_{ij},r_{ij}}$. To introduce heterogeneity across multi-resolution sub-regions,  
we added region-specific random shifts, $\bm{e}^k_{i_k}\sim \mathcal{MVN}(\bm{0},\bm{\Lambda})$, $k=1,\ldots,K$, $i_k=1,\ldots,k^2$, on $\bm{y}_{ij}$, with $i_k$ denoting the indicator for sub-region under $k\times k$ segmentation, and $i^k_{ij}$ denoting the indicator for the sub-region the $j$-th voxel in the $i$-th image belongs to under $k\times k$ segmentation.

We set the mean and covariance of the mpMRI parameters, as well as cancer prevalence in the PZ and CG, equal to their estimates from the motivating data set. 
We varied $q_{r,0}$s and $\bm{\Lambda}$ to simulate either
moderate or strong regional heterogeneity.
We also varied $\bm{\theta} = \{\sigma^2,\phi,\nu\}$ to simulate either moderate or strong correlation between voxels within each image. 
Detailed simulation settings are described in Appendix A.1 of the supplementary materials. The spatial Gaussian kernel smoothing bandwidth was selected via Cross-Validation to maximize the AUC of the base learner under each resolution. 
We conducted 5-fold Cross-Validation on 34 synthetic subjects in each simulation, and summarized classification results under each simulated scenario based on 100 simulations.
\begin{table}
\tbl{ Simulation results of binary PCa classification assuming strong regional heterogeneity. $\sigma^2$, $\phi$ and $\nu$ are the parameters for the variance, range and smoothness, respectively, of the spatial covariance among voxels. S80 and S90 denote sensitivity corresponding to 80\% and 90\% specificity, respectively. Results are reported in the form of mean value (standard error).}
{\begin{tabular}{ccccccc}\toprule
            \multirow{2}{*}{\makecell{Spatial\\Correlation}} & \multirow{2}{*}{Base Learner} & \multirow{2}{*}{Method}
		    & \multicolumn{3} {c} {Classification Results} \\
		    &&& AUC & S80 & S90 \\	    
		    \midrule
			\multirow{11}{*}{\makecell{Moderate\\$\sigma^2=4$\\$\phi=0.2$\\$\nu=0.8$}} & \multirow{3}{*}{GLM} & Baseline & .747 (.035) & .544 (.065) & .382 (.068) \\
			&& SL0 & .829 (.005) & .697 (.009) & .544 (.010)\\
			&& SL & .842 (.010) & .732 (.019) & .575 (.022)\\	
			\cmidrule{3-6}
			& \multirow{3}{*}{QDA} & Baseline & .750 (.033) & .548 (.063) & .386 (.065) \\
			&& SL0 & .836 (.005) & .710 (.009) & .560 (.011)\\
			&& SL & .847 (.010) & .744 (.019) & .589 (.022)\\			
			\cmidrule{3-6}
			& \multirow{3}{*}{RF} & Baseline & .789 (.014) & .621 (.028) & .460 (.030) \\
			&& SL0 & .820 (.005) & .685 (.006) & .532 (.006)\\
			&& SL & .852 (.006) & .751 (.013) & .593 (.016)\\
			\cmidrule{3-6}
			& \multirow{2}{*}{GLM + QDA + RF} & SL0 & .838 (.004) & .714 (.007) & .564 (.008) \\
			&& SL & .864 (.006) & .775 (.012) & .622 (.015) \\
			\midrule
			\multirow{11}{*}{\makecell{Strong\\$\sigma^2=10$\\$\phi=0.5$\\$\nu=1.5$}} & \multirow{3}{*}{GLM} & Baseline & .746 (.040) & .542 (.073) & .380 (.074) \\
			&& SL0 & .829 (.005) & .697 (.010) & .543 (.012)\\
			&& SL & .920 (.012) & .880 (.024) & .773 (.038)\\			
			\cmidrule{3-6}
			& \multirow{3}{*}{QDA} & Baseline & .748 (.039) & .544 (.075) & .382 (.075) \\
			&& SL0 & .836 (.005) & .709 (.009) & .558 (.011)\\
			&& SL & .926 (.011) & .891 (.020) & .790 (.032)\\			
			\cmidrule{3-6}
			& \multirow{3}{*}{RF} & Baseline & .790 (.018) & .621 (.035) & .460 (.038) \\
			&& SL0 & .820 (.005) & .685 (.008) & .532 (.009)\\
			&& SL & .929 (.009) & .899 (.018) & .803 (.027)\\			
			\cmidrule{3-6}
			& \multirow{2}{*}{GLM + QDA + RF} & SL0 & .838 (.004) & .714 (.008) & .564 (.009) \\
			&& SL & .945 (.005) & .927 (.009) & .844 (.017) \\		
			\bottomrule
		\end{tabular}}
\label{simu_binary_table_strong}
\end{table}
\FloatBarrier

Simulation results are summarized by the empirical mean and standard error (SE) of the AUC, S80 and S90 (sensitivity corresponding to 80\% and 90\% specificity, respectively), under moderate (Table S1 in Appendix B of the supplementary materials) and strong (Table \ref{simu_binary_table_strong}) regional heterogeneity, respectively. 
Among the three considered base learners, GLM and QDA both outperform RF and show similar classification accuracy. By comparing ``SL0'' with ``baseline'', we can observe that our proposed approach improves the AUC, S80 and S90 of the base learners by capturing the complex local heterogeneity via the multi-resolution modeling strategy, and the improvement increases as the magnitude of regional heterogeneity increases. 
The multi-resolution modeling strategy also substantially reduces the SD of AUC, S80 and S90 in all simulation settings.
By comparing ``SL'' with ``SL0'', we can observe that the intermediate spatial smoothing step further improves classification by accounting for the spatial dependency between voxels, where the improvement increases as the magnitude, scale and smoothness of the spatial correlation increase. Implementing spatial smoothing inflates the SE of AUC, S80 and S90 obtained from SL, but the SE remains smaller than that of the base learner. 
Without the intermediate spatial smoothing step, combining GLM, QDA and RF (GLM + QDA + RF) provides similar classification accuracy as the best single learner-based classifier (see ``SL0''s in Tables S1 and Table \ref{simu_binary_table_strong}). After incorporating spatial smoothing, however, GLM + QDA + RF leads to improved classification compared to the single learner-based classifiers with spatial smoothing (see ``SL''s in Tables S1 and Table \ref{simu_binary_table_strong}). The stage-one models trained under different resolutions ($k=1,2,3$) tend to have similar weights in the stage-two model, indicating important contributions of all different resolutions, which highlights the reason our proposed multi-resolution modeling strategy outperforms the conventional classifiers. 
The similar weights of different resolutions are also as expected since in our simulations the local features under different resolutions were generated from the same distribution.  
It took 2.28 hours for the ``SL'' model that combined GLM, QDA and RF to complete 100 simulations, which showed high computational efficiency of the proposed approach compared to the previously proposed Bayesian spatial models (see Jin et al., 2020 \cite{jin2020}).

\subsection{Application to patient data} \label{sec4.2}
We now illustrate the performance of the various binary classifiers on our motivating data set described in Section \ref{sec2}. Table \ref{application_binary_table} summarizes the  classification results obtained from 5-fold Cross-Validation.
When using GLM as the base learner, the proposed  multi-resolution modeling approach improves the AUC from 0.735 to 0.775, the S80 from 0.582 to 0.651, and the S90 from 0.423 to 0.514. The intermediate spatial smoothing step further improves the AUC to 0.819, the S80 to 0.728, and the S90 to 0.590.
Similar improvements can be observed when using QDA or RF as the base learner. Without the intermediate spatial smoothing step, the proposed algorithm using GLM or QDA as the base learner provides higher classification accuracy than when using RF as the base learner (AUC: 0.775 and 0.761 v.s. 0.738). But after incorporating spatial smoothing, the RF-based super learner performs better than the GLM and QDA-based learners (AUC: 0.836 v.s. 0.819 and 0.803).

\begin{table}
\tbl{Binary classification results on patient data.
The weight for each resolution was calculated as the average from 5-fold Cross-Validation.
}	
{\begin{tabular}{cccccccc}
			\toprule
		    \multirow{2}{*}{Base Learner} & \multirow{2}{*}{Method}
		    & \multicolumn{3} {c} {Classification Results}
		    & \multicolumn{3} {c} {\makecell{Weight for\\Each Resolution}}
		    \\
		    && AUC & S80 & S90 & $1\times 1$ & $2\times 2$ & $3\times 3$\\
		    \midrule
			\multirow{3}{*}{GLM}
			& Baseline & 0.735 & 0.582 & 0.423 & \rule{0.3cm}{0.4pt} & \rule{0.3cm}{0.4pt} & \rule{0.3cm}{0.4pt} \\
			& SL0 & 0.775 & 0.651 & 0.514&0.818&1.205&1.483\\
			& SL & 0.819 & 0.728 & 0.590 & 1.761 & 1.372 & 2.956\\
			\cmidrule{2-8}	    
			\multirow{3}{*}{\makecell{QDA}}
			& Baseline & 0.737 & 0.594 & 0.431 & \rule{0.3cm}{0.4pt} & \rule{0.3cm}{0.4pt} & \rule{0.3cm}{0.4pt}\\
			& SL0 & 0.761 & 0.635 & 0.490 & 1.301 & 0.757 & 0.854\\
			& SL & 0.803 & 0.696 & 0.569 & 2.274 & 1.018 &  1.811\\
			\cmidrule{2-8}
			\multirow{3}{*}{\makecell{RF}}
			& Baseline & 0.685 & 0.485 & 0.292 & \rule{0.3cm}{0.4pt} & \rule{0.3cm}{0.4pt} & \rule{0.3cm}{0.4pt}\\
			& SL0 & 0.738 & 0.563 & 0.408 & 0.780 & 0.637 & 1.029\\	
			& SL & 0.836 & 0.738 & 0.587 & 2.866 & 1.540 & 2.694\\
			\cmidrule{2-8}		
			\multirow{2}{*}{\makecell{GLM + QDA + RF}}
			& SL0 & 0.778 & 0.647 & 0.496 & \rule{0.3cm}{0.4pt} & \rule{0.3cm}{0.4pt} & \rule{0.3cm}{0.4pt}\\
			& SL & 0.825 & 0.709 & 0.543 & \rule{0.3cm}{0.4pt} & \rule{0.3cm}{0.4pt} & \rule{0.3cm}{0.4pt}\\
			\bottomrule
		\end{tabular}}
\label{application_binary_table}
\end{table}
\FloatBarrier

\begin{figure}[ht!]
\centering
\includegraphics[width=\textwidth]{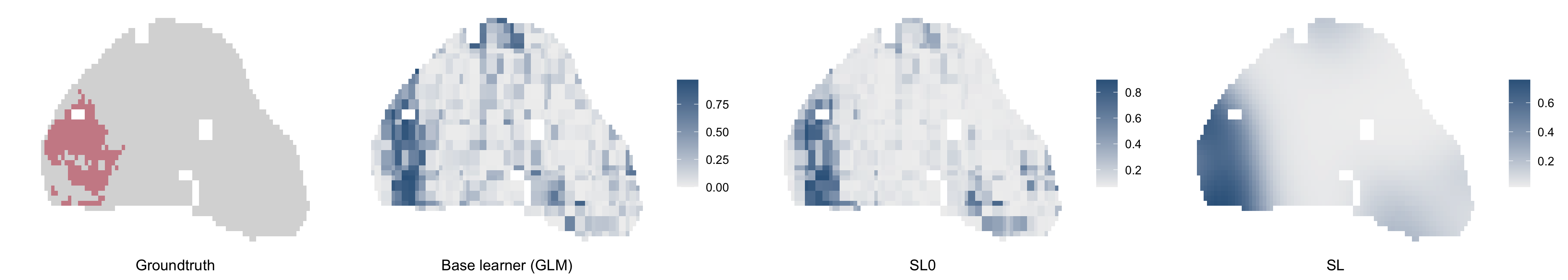}\\
\includegraphics[width=\textwidth]{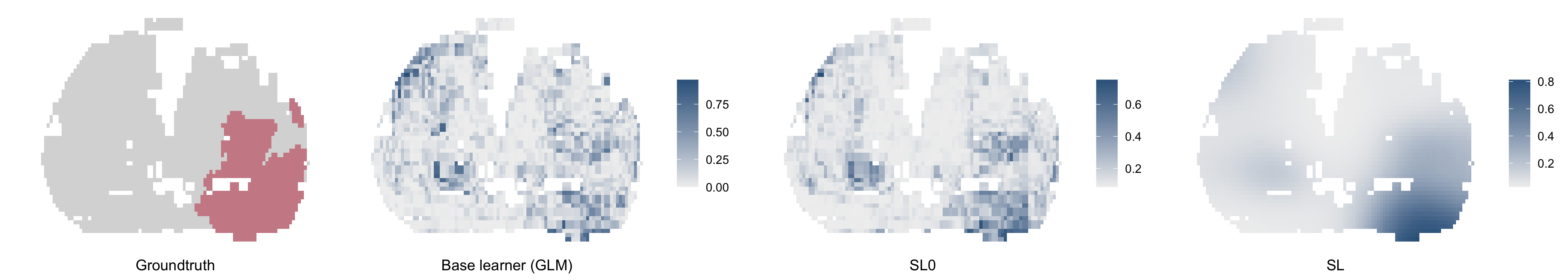}\\
\includegraphics[width=\textwidth]{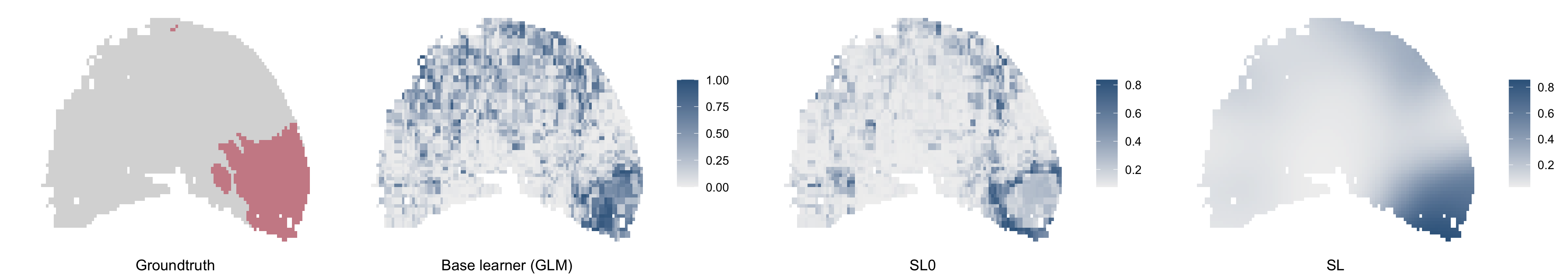}
\caption{Maps of three randomly selected prostate slices in our motivating dataset showing the groundtruth of voxel-wise PCa status (column 1, where red indicates cancer), and the GLM-based prediction of the voxel-wise cancer probabilities obtained from the base learner (GLM, column 2), the proposed algorithm without spatial smoothing (``SL0'', column 3), and the proposed algorithm with spatial smoothing (``SL''), respectively. The values in the bar of each prediction map are the predicted voxel-wise cancer probabilities.
\label{fig2.2}}
\end{figure}

An interesting finding is that, without the intermediate spatial smoothing step, combining the multi-resolution GLM, QDA and RF-based learners has an AUC slightly higher than but similar to that of the best single learner-based classifier (0.778 for GLM + QDA + RF v.s. 0.775 for GLM, 0.761 for QDA and 0.738 for RF). 
But after adding the spatial smoothing step, the RF-based super learner provides higher classification accuracy than GLM + QDA + RF (AUC: 0.836 v.s. 0.825).
This is possibly because spatial smoothing reduces variation due to spatial correlation and random noise, but after combining GLM, QDA, and RF-based learners together, 
most of the noise in the data that would otherwise be reduced by spatial smoothing have already been removed, and, as a result, 
there is less room for improvement. 
Figure \ref{fig2.2} shows the GLM-based prediction of the voxel-wise PCa probabilities for three randomly selected prostate slices in our motivating dataset. By comparing the results from SL0 with those from the baseline model (GLM), we can see that our proposed multi-resolution modeling approach substantial reduces the overly high predicted cancer probabilities for non-cancer voxels. Compared to SL0, SL continues to reduce potential false positives and produces more integrated regions of high cancer probabilities that have substantial overlaps with the true cancer regions.

\FloatBarrier

Table \ref{application_binary_table} also reports the estimated weight for each resolution 
averaged across the 5 folds of Cross-Validation, which describe the relative contribution of the models trained under different resolutions. 
For the GLM-based super learners, the local learner with the highest resolution, $3\times 3$, has the largest weight, indicating that GLM has captured more higher-resolution local features. The global QDA-based learners trained under resolution $1\times 1$ has the largest weight, indicating that QDA captures more global features in the data that are crucial for the classification. For the RF-based learners, the global learner with resolution $1\times 1$ and the local learner with resolution $3\times 3$ have higher weights than the learner with resolution $2\times 2$, indicating that more data features have been captured by RF under these two resolutions. 
Although different resolutions have different weights, the magnitude of the weights tend to be similar. 
This indicates that the base learners trained under the three different resolutions have all made substantial contributions to the classification, which explains why our multi-resolution modeling approach outperforms the global classifiers.


\section{Extension to classifying the ordinal clinical significance of PCa} \label{sec5}
The proposed method was motivated by the need for an approach that could be easily extended to tackle more complex classification/prediction problems for PCa. In this Section, we will discuss an extension of our method to classifying the ordinal outcome $G$ defined in Section \ref{sec2}, i.e., the clinical significance of PCa, which is critical for selecting appropriate treatments in clinical practice.

\subsection{Method} \label{sec5.1}
Recall that the voxel-wise indicator for clinical significance of PCa is defined based on Gleason score: $G=1$ (noncancer) if $S^{a}<3$ and $S_a + S_b < 6$, $G=2$ (clinically insignificant cancer) if $S_a + S_b = 3+3$ or $3+4$, and $G=3$ (clinically significant cancer) if otherwise. 
Here $G_{ij}$s have $Z=3$ ordered levels. In a more general situation where, for example, we would like to predict the Gleason score directly, $Z$ can have larger values. 
Implementation of the ordinal classification algorithm still follows the workflow proposed for binary classification in Section \ref{sec3.2}, but with the following modifications. 
First, the selected base learners are the classifiers that can handle ordinal outcomes, such as ordinal probit regression. 
Second, given the ordinal outcome, there are multiple choices for $\bm{\widetilde{x}}_{ij}$, i.e., the stage-one output that summarizes classification results of the multi-resolution base learners, which will also be used as the covariates in stage-two model. For example, $\bm{\widetilde{x}}_{ij}$ can be the vector of the predicted probabilities for any two of the three categories, or simply the predicted category, $\widehat{G}_{ij}$. 
Third, we change the stage-two model from probit regression to ordered probit regression:
\begin{align*}
G_{ij} = z  \text{ if }  a_{z-1} \leqslant G^{\ast}_{ij} < a_z,\text{  } z=1,2,\ldots,Z, 
\end{align*}
where $\{a_{z},z=0,1,\ldots,Z\}$ is the set of boundaries between categories, with $a_0=-\infty$ and $a_Z=\infty$.

One practical issue that adds difficulty to ordinal PCa classification is the large difference in prevalence across different PCa categories. Take the motivating data set as an example: the prevalence of noncancer ($G=1$), clinically insignificant ($G=2$) and significant ($G=3$) cancer voxels are 0.833, 0.058 and 0.109, respectively, and, as a result, detecting the less prevalent clinically insignificant and significant PCa voxels can be challenging with limited information provided by the data.   
To increase the power of detecting less prevalent PCa voxels, we consider a weighted likelihood approach for the stage-two model \cite{hu2002,agostinelli2012}: 
\begin{align}
   L^{\bm{w}}(\bm{G}|\bm{\vartheta}) = \prod_{i=1}^N\prod_{j=1}^{n_i} f(G_{ij}|\bm{\widetilde{x}}_{ij},\bm{\vartheta})^{w_{ij}(G_{ij},\bm{\widetilde{x}}_{ij})},
\end{align}
where $\bm{\vartheta}$ denotes the set of model parameters, and $w_{ij}$ is a user-defined weight for the $j$-th voxel in the $i$-th image that is a bounded differentiable non-negative function of $G_{ij}$ and $\bm{\widetilde{x}}_{ij}$. 
The standard likelihood is equivalent to the weighted likelihood with equal weights ``$W_1$": $w_{ij}=1/\sum_{i=1}^N n_i$, $\forall i,j$. An option for unequal weights is ``$W_2$'': $w_{ij}=1/(m_{G_{ij}} Z)$, 
where $m_z$ denotes the number of voxels that belong to category $z\in \{1,2,3\}$. Based on this definition, $w_{ij}$ is inversely proportional to the prevalence of the corresponding category $G_{ij}$, and therefore we up-weight the data for rare categories and down-weight the data for more prevalent categories. This definition of weight is reasonable for classification when missing aggressive disease is deemed as a bigger issue than other types of false classifications.

We use two metrics to assess the accuracy of ordinal classification: the $Z \times Z$ classification table and overall error rate (i.e., the percentage of the falsely categorized voxels). Additionally, to evaluate the classification results for each category, we define: (1) ``False Positive Rate'' for category $z$: FPR$(z)$, the percentage of the voxels in category $z$ that are falsely classified as $z'\neq z$; (2) ``False Discovery Rate'' for category $z$: FDR$(z)$, the percentage of the voxels classified as in $z$ that are actually not in $z$:
\begin{align}
    \text{FPR}(z) & = \frac{\sum_{i=1}^N\sum_{j=1}^{n_i} \mathds{I}\left(G_{ij} = z, \widehat{G}_{ij} \neq z\right)}{\sum_{i=1}^N\sum_{j=1}^{n_i} \mathds{I}(G_{ij} = z)},\nonumber\\
    \text{FDR}(z) & = \frac{\sum_{i=1}^N\sum_{j=1}^{n_i} \mathds{I}\left(G_{ij} \neq z, \widehat{G}_{ij} = z\right)}{\sum_{i=1}^N\sum_{j=1}^{n_i} \mathds{I}(\widehat{G}_{ij} = z)}.
\end{align}
Note that the definition of these FPRs and FPRs are category-specific, i.e., the ``positive'' and ``discovery'' do not necessarily mean ``cancer-positive''.

\subsection{Simulation studies} \label{sec5.2}
We conducted simulations to evaluate the performance of our proposed algorithm for classifying the ordinal clinical significance of PCa. 
Detailed simulation settings are summarized in Appendix A.2 of the supplementary materials. 
We set the between-class boundaries, $a_1$ and $a_2$, equal to the median and 70-th percentile, respectively, of the simulated $G^{\ast}_{ij}$s, i.e., we assign a high prevalence (50\%) to the noncancer voxels, a low prevalence (20\%) to the clinically significant cancer voxels, and the lowest prevalence (10\%) to the clinically insignificant cancer voxels to mimic the scenario observed in the motivating data set. 
The generating process for $\bm{w}_i$s, $G^{\ast}_{ij}$s, $q_{r,0}$s, $\bm{e}^k_{i_{k}}$s, $\bm{\delta}_i$s and $\bm{y}_{ij}$s was similar to that in Section \ref{sec4.1}, except that 
$\bm{y}_{ij}$ depended on $G_{ij}$ instead of $c_{ij}$.  

We considered different choices for the base learner, including ordered probit regression (``GLM''), QDA, and RF. For the weighted likelihood of the stage-two model, we implemented two different sets of weights previously discussed in Section \ref{sec5.1}: (1) equal weights ``$W_1$"; (2) unequal weights ``$W_2$'' that are inversely proportional to the prevalence of the corresponding cancer categories. 
Given each base learner, we applied the following models: (1) ``Baseline'': the base learner; (2) ``SL0 + $W_1$'': the proposed algorithm without spatial smoothing and with weights $W_1$; (3) ``SL + $W_1$'': the proposed algorithm with weights $W_1$;
(4) ``SL0 + $W_2$'': the proposed algorithm without spatial smoothing and with weights $W_2$; and (5) ``SL + $W_2$'': the proposed algorithm with weights $W_2$. 
Additionally, we considered combining the stage-one, multi-resolution GLM, QDA and RF (``GLM + QDA + RF'') either with or without spatial smoothing, and using either $W_1$ or $W_2$ as the weights (``SL0 + $W_1$'', ``SL + $W_1$'', ``SL0 + $W_2$'' and ``SL + $W_2$'', respectively). 
As previously discussed, there are multiple choices for  $\bm{\widetilde{x}}_{ij}$s, the stage-one output that will be used as the covariates for stage-two model. In the simulation, we used the predicted probabilities for the first two categories as $\bm{\widetilde{x}}_{ij}$s to illustrate the performance of the method. 
The tuning parameters of the selected base learners were selected via Cross-Validation to minimize the overall error rate. 
We also used Cross-Validation to select the spatial Gaussian kernel smoothing bandwidth that minimized the overall error rate under each resolution.  
In each simulation, results were summarized by 4-fold Cross-Validation on 40 subjects.
We summarized the results by the aforementioned classification table, overall error rate, as well as the FPR and FDR for each cancer category.
\begin{table}
\tbl{Simulation results for classification of the ordinal clinical significance of PCa assuming strong regional heterogeneity and strong between-voxel correlation ($\sigma^2=10$, $\phi=0.5$, $\nu=1.5$)
using GLM as the base learner.
$W_1$ and $W_2$ denote the equal weights and the weights inversely proportional to the prevalence of the corresponding cancer categories, respectively, for the weighted likelihood of the stage-two model.}
{\begin{tabular}{ccccccccc}
			\toprule
			\multirow{2}{*}{Method} & \multirow{2}{*}{\makecell{True PCa\\Category}}
		    & \multicolumn{3}{c}{Classified Category} & \multirow{2}{*}{FPR} & \multirow{3}{*}{FDR} & \multirow{2}{*}{\makecell{Overall\\Error Rate}}\\
		    && 1 & 2 & 3 & & & \\
			\midrule
			\multirow{3}{*}{Baseline} &1 & 60217 & 0 & 9822 & 0.14 & 0.38 & \multirow{3}{*}{0.42}\\
			& 2 & 17083 & 0 & 10932  &1.00 & NA & \\
			& 3 & 20475 & 0 & 21548 &0.49 & 0.49 & \\\\
			\multirow{3}{*}{SL0 + $W_1$} &1 & 61296&0&8763 & 0.13 & 0.34 & \multirow{3}{*}{0.38}\\ %
			& 2 & 15311&0&12704 &1.00 & NA & \\
			& 3 & 16359&0&25664 &0.39 & 0.46 &  \\\\
			\multirow{3}{*}{SL + $W_1$} &1 & 64001 & 2598&3440 & 0.09 & 0.1 & \multirow{3}{*}{0.27}\\ %
			& 2 & 13379&3804&10832 &0.86 & 0.60 & \\
			& 3 & 4207&2837&34979 &0.17 & 0.29 &  \\\\
			\multirow{3}{*}{SL0 + $W_2$} &1 & 48390&13488&8160 &0.31 & 0.25 & \multirow{3}{*}{0.42} \\
			& 2 & 7797&7943&12275  &0.72 & 0.73 & \\
			& 3 & 8224&8515&25284 &0.40 & 0.45 &  \\\\ 
			\multirow{3}{*}{SL + $W_2$} &1 & 54194&13919&1925 & 0.23 & 0.12 & \multirow{3}{*}{0.29}\\ %
			& 2 & 6582&13643&7790  &0.51 & 0.63 & \\
			& 3 & 1150&9477&31396 &0.25 & 0.24 &  \\
			\bottomrule
		\end{tabular}}
\label{ordinal_simu_table_glm_strong_strongsp}
\end{table}
\FloatBarrier

Table \ref{ordinal_simu_table_glm_strong_strongsp} presents simulation results assuming strong regional heterogeneity and strong between-voxel correlation using GLM as the base learner. Results assuming weaker between-voxel correlation and weaker regional heterogeneity are reported in Tables S2 and S3, respectively, in Appendix C.1 of the supplementary materials. 
The reported results are averaged across 100 simulations per scenario. By comparing SL0 + $W_1$ to Baseline, we observe that the proposed multi-resolution modeling strategy improves the classification for categories 1 and 3, with larger improvement when there is stronger regional heterogeneity. Comparing SL + $W_1$ to SL0 + $W_1$, the spatial smoothing step further improves classification of the two categories, with larger improvement under stronger spatial correlation. Overall, the number of correctly identified clinically significant cancer voxels has a large increase from Baseline to the four models using the proposed algorithm. 
We also considered using QDA or RF as the base learner, or combining multi-resolution GLM, QDA and RF, the simulation results for which are summarized in Tables S4-S9 in Appendix C.1 of the supplementary materials. 
In general, using GLM, QDA or RF as the base learner gives similar classification results, while combining multi-resolution GLM, QDA and RF together provides similar or potentially higher classification accuracy for all categories compared to any single learner-based classifiers.

One noticeable finding, the rationale for which was briefly discussed in Section \ref{sec5.1}, is that correctly identifying clinically insignificant cancer voxels is challenging.  
When using GLM as the base learner, both the baseline model and our proposed algorithm with equal weights $W_1$ cannot identify any clinically insignificant cancer voxels;
although QDA and RF can correctly identify a small proportion, the corresponding super learners with equal weights $W_1$ fail to identify any, even though better overall performance is achieved. 
Both the baseline model and our proposed algorithm using equal weights sacrifice classification of the less prevalent clinically insignificant cancer category to improve the overall classification.
Howevever, with weights $W_2$ that up-weight less prevalent cancer categories, 
the multi-resolution modeling approach substantially improves the detection of the clinically insignificant cancer voxels, which, however, comes with the price of higher FPR for noncancer voxels, although the FDR of noncancer voxels is lowered.
While our algorithm with unequal weights $W_2$ does not improve the overall ordinal classification, its enhanced ability of identifying cancerous voxels can be an appealing feature in clinical practice. The spatial smoothing step, on the other hand, has further improved classification for all categories using either $W_1$ or $W_2$ as the weights. 
Note that the challenge in distinguishing clinically insignificant cancer from the other two categories comes not only from the low prevalence of the category, but also from the fact that the differences in sample mean of the mpMRI parameters between different categories are quite small considering the large sample variance, which is possibly due to the limited sample size that is commonly observed from most of the current mpMRI studies \cite{Lemaitre2015}. As long as we only use the voxel-wise, category-specific distribution of the mpMRI parameters to distinguish between cancer categories, this issue will always exist and cannot be simply addressed by the multi-resolution modeling strategy.

\subsection{Application to patient data} \label{sec5.3}
We applied the ordinal classifiers, including Baseline, SL0 + $W_1$, SL + $W_1$, SL0 + $W_2$, and SL + $W_2$, using ordinal probit regression (``GLM'') as the base learner to our motivating data set. 
Tables \ref{ordinal_application_table_glm_pred} and \ref{ordinal_application_table_glm_cat} summarize the 4-fold Cross-Validated results using predicted probabilities of the first two categories and the predicted cancer categories, respectively, as $\bm{\widetilde{x}}_{ij}$ (the stage-one output). 
Compared to Baseline, we see that the proposed multi-resolution modeling approach with equal weights (SL0 + $W_1$) gives lower overall error rates, FPR and FDR for the clinically significant cancer voxels, lower FDR and slightly higher FPR for the noncancer voxels, and still no identified clinically insignificant cancer voxels. 
Comparing SL + $W_1$ to SL0 + $W_1$, the spatial smoothing step leads to further improvements in the aforementioned directions. 
Comparing SL + $W_1$ to SL + $W_2$, we observe that using unequal weights $W_2$ enables our proposed algorithm to correctly identify some clinically insignificant cancer voxels. 
However, it leads to higher FPR for the noncancer voxels as well, although the corresponding FDR is lowered. 
With equal weights $W_1$, using predicted probabilities and using classified categories as the stage-one output give similar results.
With unequal weights $W_2$, compared to using the classified categories as the stage-one output, using predicted probabilities tends to identify more clinically significant and insignificant cancer voxels, with lower FPR for both categories, but also higher FDR for the two categories, higher FPR for the noncancer voxels, and higher overall error rate. 

\begin{table}
\tbl{Ordinal classification results on patient data using GLM as the base learner and predicted probabilities for categories 1 and 2 as the stage-one output.}
{\begin{tabular}{ccccccccc}
			\toprule
			\multirow{2}{*}{Method} & \multirow{2}{*}{\makecell{True PCa\\Category}}
		    & \multicolumn{3}{c}{Classified Category} & \multirow{2}{*}{FPR} & \multirow{3}{*}{FDR} & \multirow{2}{*}{\makecell{Overall\\Error Rate}}\\
		    && 1 & 2 & 3 & & & \\
		    \midrule
			\multirow{3}{*}{Baseline} &1 & 88492 & 0 & 495 & 0.007 & 0.159 & \multirow{3}{*}{0.165}\\
			& 2 & 6151 & 0 & 92  &1.000 & NA & \\
			& 3 & 10930 & 0 & 777 &0.907 & 0.437 & \\ \\
			\multirow{3}{*}{SL0 + $W_1$} &1 & 87030&0&1957 & 0.022 & 0.138 & \multirow{3}{*}{0.158}\\
			& 2 & 5953&0&290  &1.000 & NA & \\
			& 3 & 8681&0&3026 &0.698 & 0.405 &  \\ \\
			\multirow{3}{*}{SL + $W_1$} &1 & 87022&0&1965 & 0.022 & 0.129 & \multirow{3}{*}{0.140}\\
			& 2 & 5951&0&292  &1.000 & NA & \\
			& 3 & 6769&0&4938 &0.598 & 0.336 &  \\ \\
			\multirow{3}{*}{SL0 + $W_2$} &1 & 65893&15149&7945 &0.241 & 0.073 & \multirow{3}{*}{0.304} \\
			& 2 & 2442&1840&1961  &0.718 & 0.905 & \\
			& 3 & 2949&2100&6658 &0.432 & 0.575 &  \\ \\
			\multirow{3}{*}{SL + $W_2$} &1 & 63726&18912&6349 & 0.253 & 0.063 & \multirow{3}{*}{0.315}\\
			& 2 & 2317&2618&1308  &0.661 & 0.901 & \\
			& 3 & 1885&2863&6959 &0.401 & 0.519 &  \\ 
			\bottomrule
		\end{tabular}}
\label{ordinal_application_table_glm_pred}
\end{table}
\FloatBarrier

\begin{table}
\tbl{Ordinal classification results on patient data using GLM as the base learner and predicted cancer categories as the stage-one output.}
{\begin{tabular}{cccccccccc}
			\toprule
			\multirow{2}{*}{Method} & \multirow{2}{*}{\makecell{True PCa\\Category}}
		    & \multicolumn{3}{c}{Classified Category} & \multirow{2}{*}{FPR} & \multirow{3}{*}{FDR} & \multirow{2}{*}{\makecell{Overall\\Error Rate}}\\
		    && 1 & 2 & 3 & & & \\
		    \midrule
			\multirow{3}{*}{Baseline} &1 & 88492 &0&495 & 0.006 & 0.162 & \multirow{3}{*}{0.165}\\
			& 2 & 6151&0&92  &1.000 & NA & \\
			& 3 & 10930&0&777 &0.934 & 0.430 &  \\ \\
			\multirow{3}{*}{SL0 + $W_1$} &1 & 87850 &0&1137 & 0.013 & 0.151 & \multirow{3}{*}{0.159}\\
			& 2 & 6083&0&160  &1.000 & NA & \\
			& 3 & 9583&0&2124 &0.819 & 0.379 &  \\ \\
			\multirow{3}{*}{SL + $W_1$} &1 & 87865&0&1122 & 0.013 & 0.138 & \multirow{3}{*}{0.143}\\
			& 2 & 6173&0&70  &1.000 & NA & \\
			& 3 & 7874&0&3833 &0.673 & 0.237 &  \\\\ 
			\multirow{3}{*}{SL0 + $W_2$} &1 & 86039&0&2948 & 0.033 & 0.131 & \multirow{3}{*}{0.155}\\
			& 2 & 5524&0&719  &1.000 & NA & \\
			& 3 & 7437&0&4270 &0.635 & 0.462 &  \\ \\
			\multirow{3}{*}{SL + $W_2$} &1 & 77261&8268&3458 &0.132 & 0.081 & \multirow{3}{*}{0.206} \\
			& 2 & 4231&1418&594  &0.773 & 0.887 & \\
			& 3 & 2576&2862&6269 &0.465 & 0.393 &  \\ 
			\bottomrule
		\end{tabular}}
\label{ordinal_application_table_glm_cat}
\end{table}
\FloatBarrier

We also considered using QDA or RF as the base learner, or combining the multi-resolution GLM, QDA and RF. Results are summarized in Tables S10-S12 in the Online Appendix C.2 of the supplementary materials. 
Overall, using GLM, QDA or RF as the base learner gives similar results, while combining GLM, QDA and RF together improves the detection of both clinically insignificant and significant PCa voxels, but decreased FPR for noncancer and higher overall error rate.

\section{Discussion} \label{sec6}
We propose a novel classification algorithm for the voxel-wise detection and grading of PCa using high-resolution mpMRI data. 
The main feature of the algorithm is the multi-resolution modeling strategy, which accounts for regional heterogeneity in the mpMRI data by averaging over global and local classifiers trained at different resolutions via super learner. 
This multi-resolution modeling strategy provides a flexible and easily implementable approach to capturing both global and local features of mpMRI, addressing limitations of the Bayesian hierarchical models recently proposed for voxel-wise classification of binary PCa status. 
We have demonstrated the advantages of our algorithm over the conventional classification methods such as GLM, QDA and RF on both synthetic mpMRI datasets and our motivating patient data.  
The proposed algorithm provides a flexible modeling framework where any classifier, including the black-box machine learning algorithms, can be implemented as the base learners. 
This indicates that the algorithm can combine multiple types of classifiers to further improve classification, and can be continuously enhanced and updated once more powerful algorithms become available. We believe that in practice the resulting PCa classifiers can help provide a better guidance for PCa diagnosis and decisions for appropriate treatment. The proposed multi-resolution modeling approach can also be applied to other types of imaging data to capture the potential local, heterogeneous features that differ from the global, homogeneous patterns in the images.

An important feature of the proposed method is that it can be extended to deal with more complex classification problems. As an illustration, we extended the binary classification algorithm to classify the ordinal clinical significance of PCa based on Gleason grading system. Improvements in detecting clinically significant PCa were shown by both simulations and application to real patient data, which means that potentially enhanced guidance for clinical treatment can be achieved. 
We propose to use weighted likelihood for the stage-two model in our algorithm to improve classification of the less prevalent cancer categories. 
Depending on the primary goal, the weights can be adjusted to enhance detection of some specific cancer categories. 


Our study has limitations. While we observed improvements in performance as the result of our proposed classifier, the classification accuracy of the trained classifiers is still relatively low. Although we can adjust the parameter cutoffs to control the ``false positive rate'' of each PCa category according to our needs, the overall false classification rate, especially for the ordinal clinical significance outcome, is higher than desired. 
This is a general limitation of voxel-wise methods for predicting PCa significance, which rely on the voxel-level, cancer-category-specific means and covariances of the mpMRI parameters. As shown in Supplementary Figures S1 - S3, these quantities have small differences across PCa categories that are hard to detect given the currently limited sample size for most of the mpMRI studies. 
Theoretically, we could combine multiple studies to obtain improved classifiers. combining datasets is challenging due to variability in the data collection protocols with respect to the MRI scanner, mpMRI parameters calculated, or/and resolution of the generated mpMRI images. 
To achieve fundamental improvement in voxel-wise PCa classification, novel modeling approaches should be developed to replace the existing ones that are mainly based on the cancer-category-specific mean of the mpMRI parameters. For example, quantile regressions\cite{koenker2001quantile}, which show advantages over mean regressions in predicting the relationship between variables when there is only weak association between the means of the variables, can be considered. 
Additionally, a two-step, region-wise modeling approach can be considered, where we first locate candidate cancer lesions based on voxel-wise classification results, then determine their aggressiveness by their similarity to areas of each cancer category in terms of lesion-wise mpMRI distributions. 

In this paper, we describe our model development using only one 2D prostate slice per patient due to the current technical difficulties of co-registering 3D mpMRI and outcome data. Once the 3D co-registered data is available, we will train the proposed classifiers on the 3D data simply by extending the multi-resolution modeling procedure from 2D to 3D, which could further improve classification. 
Our super learner algorithm can also be extended to incorporate more complex model assumptions.
We currently assume that the weight for each resolution is constant across the whole prostate gland. 
Alternately, we can allow the weights to vary by the location in a prostate.  Additionally, instead of a post-hoc procedure, the spatial smoothing process can be incorporated in the modeling process, which, however, may diminish scalability of the resulting classifier and thus requires careful investigation. 
\\\\

\section*{Funding}
This work was supported by NCI R01 CA155268, NCI P30 CA077598, NIBIB P41 EB027061, and the Assistant Secretary of Defense for Health affairs, through the Prostate Cancer Research Program under Award No. W81XWH-15-1-0478. 
Opinions, interpretations, conclusions, and recommendations are those of the author and are not necessarily endorsed by the Department of Defense.

\section*{Code Availability}
The R code for implementing of the proposed classification algorithms are available at \url{https://github.com/Jin93/Multi-Resolution-SL}.

\bibliographystyle{tfs} 

\newpage
\textbf{Figure Captions}
\begin{enumerate}
    \item[] \textbf{Figure 1.} Annotated voxel-wise cancer status of an example prostate slice. Red and green indicate annotated cancer and noncancer voxels, respectively. The yellow dashed curve divides the prostate gland into peripheral zone (PZ, the area inside the curve) and central gland (CG, the area outside the curve).\\\\
    
    \item[] \textbf{Figure 2.} Region segmentation for three example prostates ($p_1$, $p_2$ and $p_3$) under resolutions $k\times k$, $k\in\{1,2,3\}$. \\\\
    
    \item[] \textbf{Figure 3.} Workflow of the proposed classification algorithm for voxel-wise binary PCa status, $c_{ij}$s. ${\Psi}_{k}$ denotes the base learner under resolution $k$, 
    $\hat{\Psi}_{k,l}^v$ denotes the learner trained in the $l$-th sub-region, $A_{k,l}$, under resolution $k$ based on the $v$-th split of Cross-Validation, and $\hat{\Psi}_{k}^v$ summarizes the results from $\hat{\Psi}_{k,l}^v$, $l=1,2,\ldots,k^2$. The total number of resolutions used for prostate segmentation is set to $K=3$ for illustration. \\\\
    
        \item[] \textbf{Figure 4.} Maps of three randomly selected prostate slices in our motivating dataset showing the groundtruth of voxel-wise PCa status (column 1, where red indicates cancer), and the GLM-based prediction of the voxel-wise cancer probabilities obtained from the base learner (GLM, column 2), the proposed algorithm without spatial smoothing (``SL0'', column 3), and the proposed algorithm with spatial smoothing (``SL''), respectively. The values in the bar of each prediction map are the predicted voxel-wise cancer probabilities.
\end{enumerate}

\vspace{20pt}
Manuscript word count: 7041 (excluding Funding, Coding Availability, References and Figure Captions). 

\vspace{20pt}
Please note that an earlier version of the manuscript was posted on the preprint server arXiv \url{https://arxiv.org/abs/2007.00816}.

\end{document}


\maketitle
\label{firstpage}

\section*{Appendix A: Detailed simulation settings}\label{c.1}
\subsection*{A.1 Simulation settings for binary PCa classification}
As described in Section 4.1 of the main manuscript, the simulated data sets were generated according to model (\ref{eq.s1}):
\begin{align}
\bm{w}_i \sim 
\mathcal{MVN}(\bm{0},\bm{C}(\bm{S}_i,\bm{S}_i| & \bm{\theta})), \text{  }
c^{\ast}_{ij} \sim N(q_{r_{ij},0} + w_{ij},1),\nonumber\\
c_{ij} = I (c^{\ast}_{ij}>0),\text{  } &
\bm{e}^k_{i_k} \sim \mathcal{MVN} (\bm{0},\bm{\Lambda}), \nonumber\\
\bm{y}_{ij} \stackrel{ind}{\sim} \mathcal{MVN} ( \bm{\mu}_{c_{ij},r_{ij}} & +   \sum_{k=1}^K\bm{e}^k_{i_{ij}^k} + \bm{\delta}_i, \bm{\Gamma}_{c_{ij},r_{ij}}).
\label{eq.s1}
\end{align}

The model parameters, including the means, $\{\bm{\mu}_{c,r},c=0,1,r=0,1\}$,  within-patient covariance, $\{\bm{\Gamma}_{c_{ij},r_{ij}},c,r\in \{0,1\}\}$, of the mpMRI parameters, and probit of the cancer prevalence in the PZ and CG, $q_{r_{ij},0}$, were set based on the estimates from the real data. 
$\bm{\mu}_{c,r}$, $c,r\in \{0,1\}$, was set equal to the estimates from the motivating data set. 
$\bm{\Gamma}_{c,r}$, $c,r\in \{0,1\}$, was set equal to 1/1.5 times the estimates from the motivating data set.
We varied the values for $q_{r,0}$, $r=0,1$, and $\bm{\Lambda}$ to simulate regional heterogeneity of different magnitudes. 
For moderate regional heterogeneity, we set the cancer prevalence in the PZ and CG to $0.4$ and $0.2$, respectively, and $\bm{\Lambda}$ equal to a $d\times d$ diagonal matrix ($d=4$) with diagonal entries 5, 0.18, 0.18, and 0.18. 
For strong regional heterogeneity, we set the cancer prevalence in the PZ and CG to $0.55$ and $0.15$, respectively, and $\bm{\Lambda}$ equal to a $d\times d$ diagonal matrix with diagonal entries 10, 0.36, 0.36, and 0.36. We also varied $\bm{\theta} = \{\sigma^2,\phi,\nu\}$ to simulate different spatial correlation structures:
for moderate spatial correlation, we set $\sigma^2$ (spatial variance) equal to 4, $\phi$ (range parameter, larger $\phi$ indicates larger-scale correlation) equal to 0.2, and $\nu$ (smoothness parameter, smaller $\nu$ indicates larger differentiability) equal to 0.8; for strong spatial correlation, we set $\sigma^2=10$, $\phi=0.5$, and $\nu=1.5$.

\subsection*{A.2 Simulation settings for classification of ordinal PCa significance}
Regarding the simulation studies for classification of the ordinal clinical significance of PCa, the shapes of the simulated prostate images were still selected with replacement from the images in the motivating data set. 
Voxel-wise cancer status and mpMRI parameters were simulated according to model (\ref{simu-model-paper3-category}). 
We set the between-class boundaries $a_1$ and $a_2$ equal to the median and 70-th percentile, respectively, of the simulated $G^*_{ij}$'s, so as to assign a high prevalence (50\%) to the noncancer voxels, a low prevalence (20\%) to the clinically significant cancer voxels, and the lowest prevalence (10\%) to the clinically insignificant cancer voxels.
The generating process for $\bm{w}_i$, $G^*_{ij}$'s, $q_{r,0}$'s, $\bm{e}^k_{i_{k}}$'s, $\bm{\delta}_i$'s and $\bm{y}_{ij}$'s was similar to the one in Section 4.1 of the main manuscript, except that $\bm{y}_{ij}$ was assumed to vary by $G_{ij}$ instead of $c_{ij}$. 
\begin{align}
\bm{w}_i \sim
\mathcal{MVN} (\bm{0} , \bm{C}(\bm{S}_i,\bm{S}_i|\bm{\theta})) & , \text{  }
G^*_{ij} \sim N(q_{r_{ij},0}  + w_{ij},1), \nonumber\\ 
G_{ij} = I (a_{z-1}\leqslant G^*_{ij}< a_{z} ) & , \text{  }
\bm{e}^k_{i_{k}} \sim \mathcal{MVN}(\bm{0},\bm{\Lambda}),\nonumber\\
\bm{y}_{ij} \stackrel{ind}{\sim} \mathcal{MVN} ( \bm{\mu}_{G_{ij},r_{ij}} + & \sum_{k=1}^K\bm{e}^k_{i^k_{ij}} + \bm{\delta}_i, \bm{\Gamma}_{G_{ij},r_{ij}}).
\label{simu-model-paper3-category}
\end{align}

\newpage
\subsection*{A.2 Data distribution across different categories of PCa}
\begin{figure}[ht!]
\centering
   \includegraphics[width=0.9\textwidth]{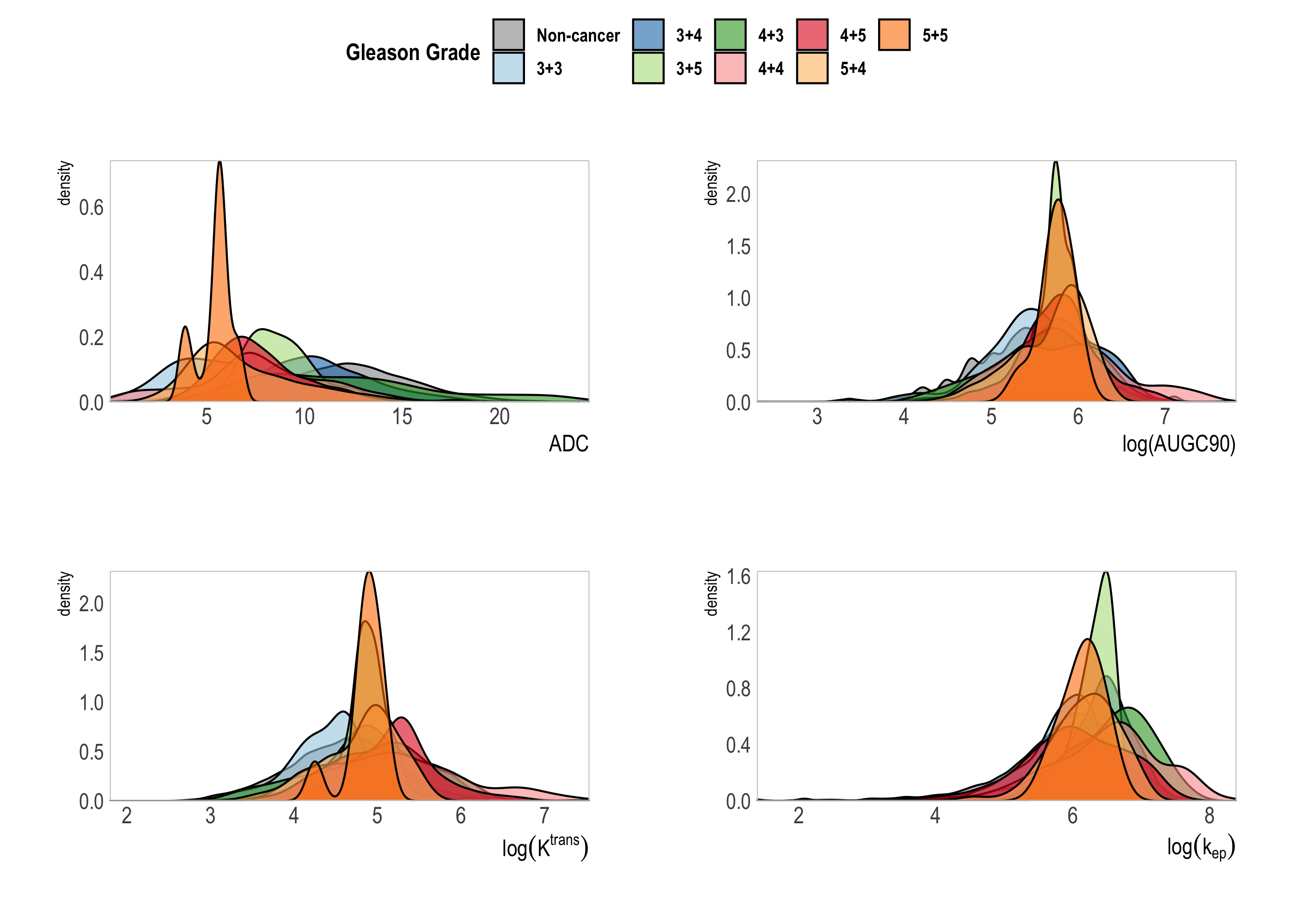}
\caption*{Figure S1. Voxel-wise distribution of the various mpMRI parameters across Gleason grade groups observed from our motivating dataset.
}
\label{fig.segmentation}
\end{figure}
\FloatBarrier

\begin{figure}[ht!]
\centering
   \includegraphics[width=0.7\textwidth]{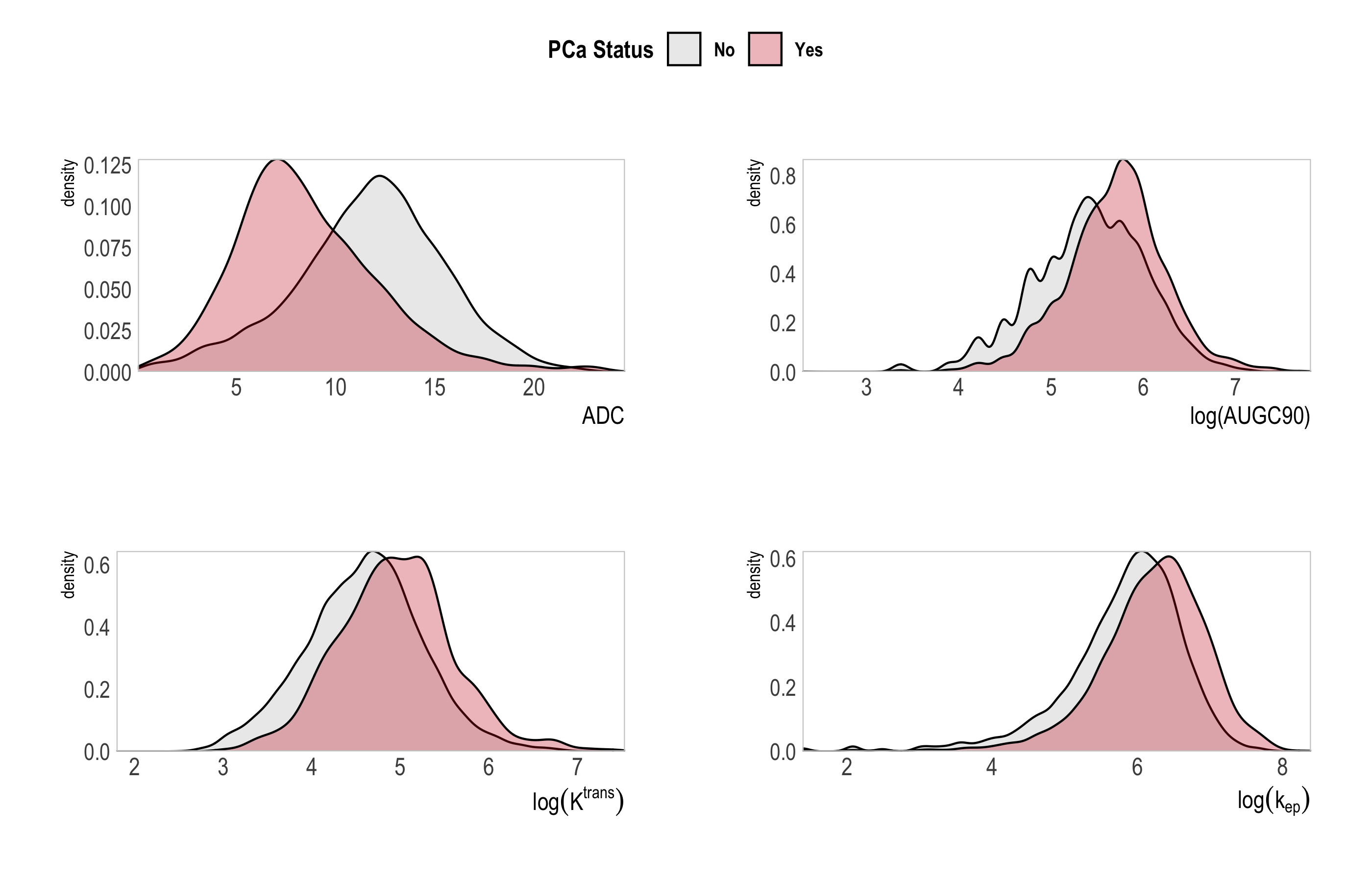}
\caption*{Figure S2. Voxel-wise distribution of the various mpMRI parameters separately for cancer and non-cancer voxels in our motivating dataset.
}
\end{figure}
\FloatBarrier

\begin{figure}[ht!]
\centering
   \includegraphics[width=0.7\textwidth]{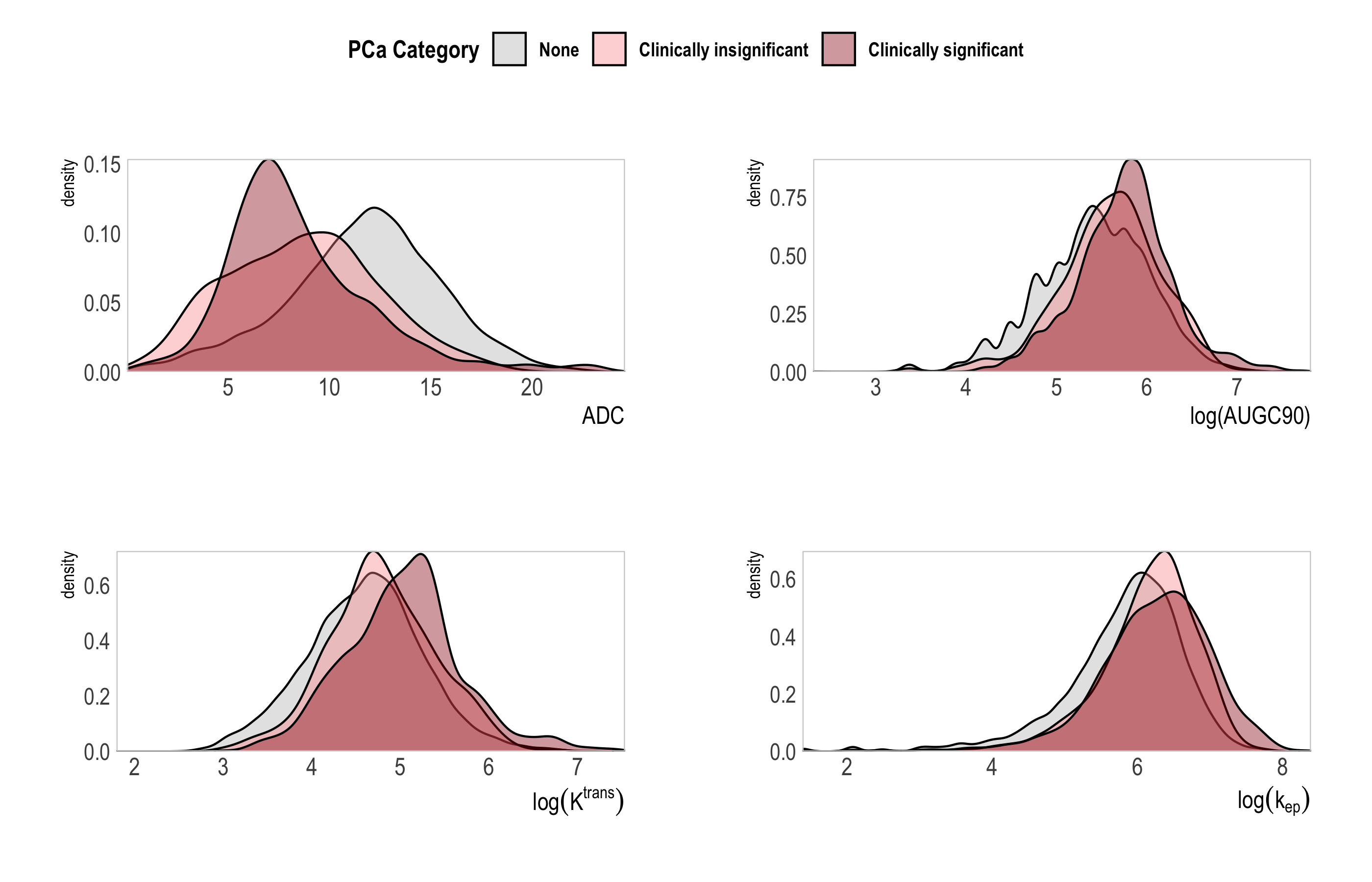}
\caption*{Figure S3. Voxel-wise distribution of the various mpMRI parameters of the voxels under different levels of clinical significance of PCa in our motivating dataset. The sample prevalences of clinically insignificant PCa voxels and clinically significant PCa voxels are 0.167.
}
\end{figure}
\FloatBarrier

\newpage
\subsection*{A.3 Examples of synthetic mpMRI and PCa maps}
\begin{figure}[ht!]
\centering
   \includegraphics[width=\textwidth]{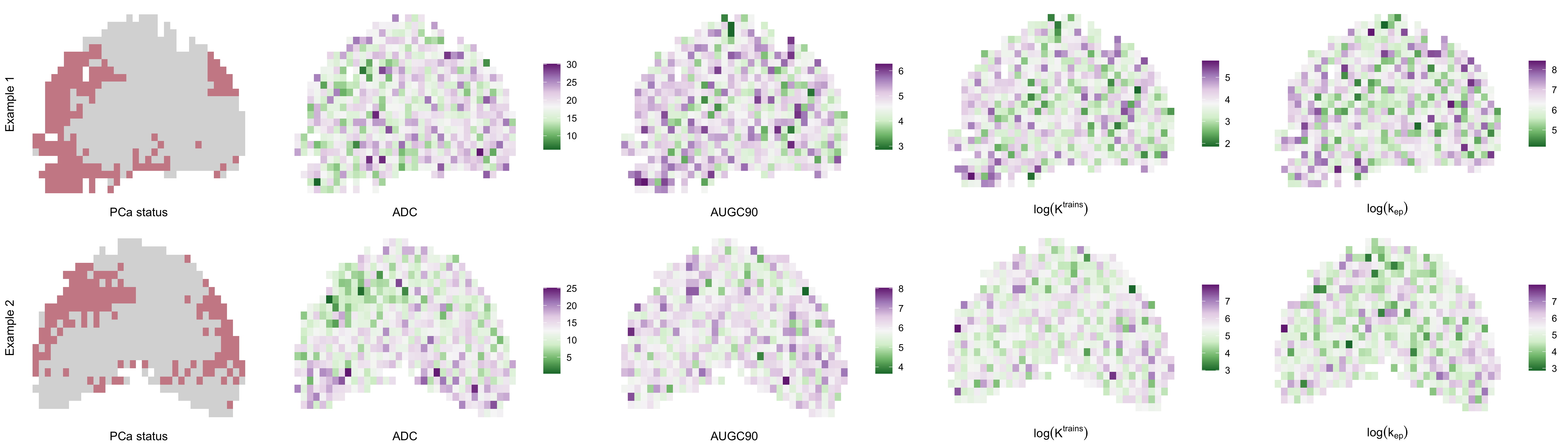}
\caption*{Figure S4. Example maps of voxel-wise PCa status (column 1) and mpMRI parameters (columns 2-5) for two simulated prostate slices under moderate regional heterogeneity and moderate spatial correlation (see Appendix A.1 for details). In the PCa outcome maps, grey and pink indicate non-cancer and cancer, respectively.
}
\label{fig.a3}
\end{figure}
\FloatBarrier

\section*{Appendix B: Additional simulation results of binary PCa classification}\label{b.0}
\begin{table}
\caption*{Table S1. Simulation results of binary PCa classification under moderate regional heterogeneity.}
\centering
{\begin{tabular}{ccccccc}\toprule
			\multirow{2}{*}{\makecell{Spatial\\Correlation}} & \multirow{2}{*}{Base Learner} & \multirow{2}{*}{Method}
		    & \multicolumn{3} {c} {Classification Results} \\
		    &&& AUC & S80 & S90 \\	    
		    \midrule
			\multirow{11}{*}{\makecell{Moderate\\$\sigma^2=4$\\$\phi=0.2$\\$\nu=0.8$}} & \multirow{3}{*}{GLM} & Baseline & .777 (.027) & .599 (.051) & .438 (.054) \\
			&& SL0 & .833 (.004) & .705 (.008) & .553 (.009)\\
			&& SL & .848 (.009) & .743 (.066) & .586 (.022)\\	
			\cmidrule{3-6}
			& \multirow{3}{*}{QDA} & Baseline & .780 (.026) & .603 (.050) & .442 (.053) \\
			&& SL0 & .840 (.004) & .717 (.008) & .567 (.009)\\
			&& SL & .854 (.009) & .754 (.018) & .598 (.021)\\
			\cmidrule{3-6}
			& \multirow{3}{*}{RF} & Baseline & .788 (.016) & .616 (.031) & .457 (.033) \\
			&& SL0 & .822 (.004) & .688 (.007) & .536 (.007)\\
			&& SL & .855 (.007) & .757 (.013) & .601 (.017)\\	
			\cmidrule{3-6}
			& \multirow{2}{*}{GLM + QDA + RF} & SL0 & .841 (.004) & .719 (.007) & .570 (.008) \\
			&& SL & .862 (.006) & .772 (.012) & .618 (.015) \\
			\midrule
			\multirow{11}{*}{\makecell{Strong\\$\sigma^2=10$\\$\phi=0.5$\\$\nu=1.5$}} & \multirow{3}{*}{GLM} & Baseline & .778 (.034) & .600 (.063) & .439 (.065) \\
			&& SL0 & .832 (.006) & .704 (.011) & .551 (.013) \\
			&& SL & .927 (.008) & .893 (.016) & .790 (.026) \\	
			\cmidrule{3-6}
			& \multirow{3}{*}{QDA} & Baseline & .778 (.034) & .600 (.063) & .439 (.064) \\
			&& SL0 & .839 (.006) & .715 (.010) & .566 (.012)\\
			&& SL & .930 (.008) & .899 (.016) & .799 (.025)\\
			\cmidrule{3-6}
			& \multirow{3}{*}{RF} & Baseline & .788 (.018) & .616 (.035) & .456 (.037) \\
			&& SL0 & .822 (.005) & .688 (.009) & .536 (.010)\\
			&& SL & .932 (.007) & .903 (.013) & .805 (.023)\\	
			\cmidrule{3-6}		
			& \multirow{2}{*}{GLM + QDA + RF} & SL0 & .840 (.005) & .718 (.009) & .569 (.011) \\
			&& SL & .942 (.005) & .921 (.010) & .833 (.017)\\
		    \bottomrule
		\end{tabular}}
\label{simu_binary_table_moderate}
\end{table}
\FloatBarrier

\section*{Appendix C: Additional results on the ordinal PCa outcome}\label{c.2}
\subsection*{C.1 Simulation studies}
Table S2 reports simulation results of models Baseline, SL0 + $W_1$, SL + $W_1$, SL0 + $W_2$ and SL + $W_2$ assuming 
more moderate between-voxel correlation than that in Tables 3 using GLM as the base learner. Table S3 reports the simulation results assuming weaker regional heterogeneity than that in Tables 3 using GLM as the base learner. 
Tables S4 and S5 report simulation results assuming moderate and strong regional heterogeneity, respectively, using QDA as the base learner. 
Tables S6 and S7 report simulation results assuming that RF is the base learner.
Tables S8 and S9 report simulation results when combining the multi-resolution GLM, QDA and RF-based learners.
\begin{table}
\caption*{Table S2. Simulation results for classification of the ordinal clinical significance of PCa assuming moderate between-voxel correlation ($\sigma^2=4$, $\phi=0.2$, $\nu=0.8$) and strong regional heterogeneity 
using GLM as the base learner.} 
\centering
\begin{tabular}{ccccccccc}
			\toprule
			\multirow{2}{*}{Method} & \multirow{2}{*}{\makecell{True PCa\\Category}}
		    & \multicolumn{3}{c}{Classified Category} & \multirow{2}{*}{FPR} & \multirow{3}{*}{FDR} & \multirow{2}{*}{\makecell{Overall\\Error Rate}}\\
		    && 1 & 2 & 3 & & & \\
		    \midrule
			\multirow{3}{*}{Baseline} &1 & 59491 & 0 & 9485 & 0.14 & 0.39 & \multirow{3}{*}{0.42}\\
			& 2 & 16928 & 0 & 10662  &1.00 & NA & \\
			& 3 & 20578 & 0 & 20808 &0.50 & 0.49 & \\\\
			\multirow{3}{*}{SL0 + $W_1$} &1 & 60466&0&8511 & 0.12 & 0.34 & \multirow{3}{*}{0.38}\\
			& 2 & 14879&0&12711 &1.00 & NA & \\
			& 3 & 15913&0&25473 &0.38 & 0.45 &  \\ \\
			\multirow{3}{*}{SL + $W_1$} &1 & 61569 & 0 & 7408 & 0.11 & 0.32 & \multirow{3}{*}{0.34}\\ %
			& 2 & 16601 & 0 & 10989  &1.00 & NA & \\
			& 3 & 11735 & 0 & 29651 &0.28 & 0.38 &  \\ \\
			\multirow{3}{*}{SL0 + $W_2$} &1 & 47771&13411&7794 &0.31 & 0.25 & \multirow{3}{*}{0.42} \\
			& 2 & 7544&7940&12106 &0.71 & 0.73 & \\
			& 3 & 7994&8589&24803 &0.40 & 0.45 &  \\\\ 
			\multirow{3}{*}{SL + $W_2$} &1 & 48110&15479&5388 & 0.30 & 0.21 & \multirow{3}{*}{0.39}\\ %
			& 2 & 8677&9910&9004  &0.64 & 0.72 & \\
			& 3 & 3990&10586&26810 &0.35 & 0.35 &  \\
			\bottomrule
		\end{tabular}
\label{ordinal_simu_table_glm_strong_modsp}
\end{table}
\FloatBarrier

\begin{table}[htb]
\footnotesize{}
\caption*{\label{ordinal_simu_table_glm_mod}
Table S3. Simulation results for classification of the ordinal clinical significance of PCa assuming moderate regional heterogeneity using GLM as the base learner.
}
\centering
    \begin{tabular}{cccccccccc}
			\toprule 
			\multirow{2}{*}{\makecell{Spatial\\Correlation}} & \multirow{2}{*}{Method} & \multirow{2}{*}{\makecell{True PCa\\Category}}
		    & \multicolumn{3}{c}{Classified Category} & \multirow{2}{*}{FPR} & \multirow{3}{*}{FDR} & \multirow{2}{*}{\makecell{Overall\\Error Rate}}\\
		    &&& 1 & 2 & 3 & & & \\
		    \midrule
			\multirow{20}{*}{\makecell{Moderate\\$\sigma^2=4$\\$\phi=0.2$\\$\nu=0.8$}}
			&\multirow{3}{*}{Baseline} &1 & 60524 & 0 & 9120 & 0.13 & 0.36 & \multirow{3}{*}{0.40}\\ %
			&& 2 & 16293 & 0 & 11564  &1.00 & NA & \\
			&& 3 & 18495 & 0 & 23292 &0.44 & 0.47 & \\\\ 
			& \multirow{3}{*}{SL0 + $W_1$} &1 & 61149&0&8495 & 0.12 & 0.33 & \multirow{3}{*}{0.37}\\ %
			&& 2 & 15064&0&12793 &1.00 & NA & \\
			&& 3 & 15673&0&26114 &0.38 & 0.45 &  \\\\ 
			& \multirow{3}{*}{SL + $W_1$} &1 & 62275 & 0 & 7370 & 0.11 & 0.31 & \multirow{3}{*}{0.33}\\
			&& 2 & 16593 & 0 & 11265  &1.00 & NA & \\
			&& 3 & 10920 & 0 & 30867 &0.26 & 0.38 &  \\\\ 
			& \multirow{3}{*}{SL0 + $W_2$} &1 & 48481&13529&7635 &0.30 & 0.24 & \multirow{3}{*}{0.41} \\ %
			&& 2 & 7616&8199&12042 &0.71 & 0.73 & \\
			&& 3 & 7914&8589&25284 &0.39 & 0.44 &  \\\\ 
			& \multirow{3}{*}{SL + $W_2$} &1 & 49119&15442&5083 & 0.29 & 0.20 & \multirow{3}{*}{0.37}\\
			&& 2 & 8465&10511&8882  &0.62 & 0.71 & \\
			&& 3 & 3528&10717&27542 &0.34 & 0.34 &  \\
			\midrule
			\multirow{20}{*}{\makecell{Strong\\$\sigma^2=10$\\$\phi=0.5$\\$\nu=1.5$}}
			&\multirow{3}{*}{Baseline} &1 & 60165 & 0 & 9323 & 0.13 & 0.37 & \multirow{3}{*}{0.40}\\ %
			&& 2 & 16341 & 0 & 11454  &1.00 & NA & \\
			&& 3 & 18520 & 0 & 23174 &0.44 & 0.47 & \\\\ 
			& \multirow{3}{*}{SL0 + $W_1$} &1 & 60923&0&8561 & 0.12 & 0.34 & \multirow{3}{*}{0.38}\\ %
			&& 2 & 15149&0&12646 &1.00 & NA & \\
			&& 3 & 15760&0&25933 &0.38 & 0.45 &  \\\\ 
			& \multirow{3}{*}{SL + $W_1$} &1 & 63432 & 3336&2720 & 0.09 & 0.20 & \multirow{3}{*}{0.25}\\ %
			&& 2 & 12317&5206&10272  &0.81 & 0.57 & \\
			&& 3 & 3160&3572&34962 &0.16 & 0.27 &  \\\\ 
			& \multirow{3}{*}{SL0 + $W_2$} &1 & 48395&13376&7718 &0.30 & 0.24 & \multirow{3}{*}{0.41} \\ %
			&& 2 & 7695&8090&12010  &0.71 & 0.73 & \\
			&& 3 & 7901&8472&25321 &0.39 & 0.44 &  \\\\ 
			& \multirow{3}{*}{SL + $W_2$} &1 & 54941&12961&1587 & 0.21 & 0.11 & \multirow{3}{*}{0.27}\\ %
			&& 2 & 6216&14062&7516  &0.49 & 0.61 & \\
			&& 3 & 916&8999&31778 &0.24 & 0.22 &  \\ 
			\bottomrule
		\end{tabular}
\end{table}
\FloatBarrier

\begin{table}[htb]
\footnotesize{}
\begin{center}
\caption*{Table S4. Simulation results for classification of the ordinal clinical significance of PCa assuming moderate regional heterogeneity using QDA as the base learner.
\label{ordinal_simu_table_qda_mod}}
    \begin{threeparttable} 	
    \begin{tabular}{cccccccccc}
			\toprule
			\multirow{2}{*}{\makecell{Spatial\\Correlation}} & \multirow{2}{*}{Method} & \multirow{2}{*}{\makecell{True PCa\\Category}}
		    & \multicolumn{3}{c}{Classified Category} & \multirow{2}{*}{FPR} & \multirow{3}{*}{FDR} & \multirow{2}{*}{\makecell{Overall\\Error Rate}}\\
		    &&& 1 & 2 & 3 & & & \\
		    \midrule
			\multirow{20}{*}{\makecell{Moderate\\$\sigma^2=4$\\$\phi=0.2$\\$\nu=0.8$}}& \multirow{3}{*}{Baseline} &1 & 59145 & 1870 & 8630 & 0.15 & 0.35 & \multirow{3}{*}{0.39}\\ %
			&& 2 & 14948 & 3272 & 9637  &0.88 & 0.52 & \\
			&& 3 & 17053 & 1691 & 23043 &0.45 & 0.44 & \\\\ 
			& \multirow{3}{*}{SL0 + $W_1$} &1 & 61215 &0&8429 & 0.12 & 0.33 & \multirow{3}{*}{0.36}\\ %
			&& 2 & 16169 & 0 & 11688 &1.00 & NA & \\
			&& 3 & 14361 & 0 & 27425 &0.34 & 0.42 &  \\\\	
			& \multirow{3}{*}{SL + $W_1$} &1 & 62875 & 0 & 6770 & 0.10 & 0.30 & \multirow{3}{*}{0.32}\\ %
			&& 2 & 16572 & 0 & 11285  &1.00 & NA & \\
			&& 3 & 9894 & 0 & 31893 &0.24 & 0.36 &  \\\\ 
			& \multirow{3}{*}{SL0 + $W_2$} &1 & 49249&13657&6738 &0.29 & 0.24 & \multirow{3}{*}{0.39} \\ %
			&& 2 & 8474&10047&9337 &0.64 & 0.69 & \\
			&& 3 & 7075&9091&25620 &0.39 & 0.39 &  \\\\ 
			& \multirow{3}{*}{SL + $W_2$} &1 & 49691&15652&4301 & 0.29 & 0.19 & \multirow{3}{*}{0.36}\\ %
			&& 2 & 8331 & 11013 & 8513  &0.60 & 0.70 & \\
			&& 3 & 3180 & 10332 & 28274 &0.32 & 0.31 &  \\
			\midrule
			\multirow{20}{*}{\makecell{Strong\\$\sigma^2=10$\\$\phi=0.5$\\$\nu=1.5$}}
			&\multirow{3}{*}{Baseline} &1 & 58734 & 1898 & 8856 & 0.15 & 0.35 & \multirow{3}{*}{0.39}\\ %
			&& 2 & 14991 & 3259 & 9545  &0.88 & 0.53 & \\
			&& 3 & 17068 & 1762 & 22862 &0.45 & 0.45 & \\\\ 
			& \multirow{3}{*}{SL0 + $W_1$} &1 & 60964 &0&8525 & 0.12 & 0.33 & \multirow{3}{*}{0.37}\\ %
			&& 2 & 16127 & 0 & 11668 &1.00 & NA & \\
			&& 3 & 14424 & 0 & 27270 &0.35 & 0.43 &  \\\\	
			& \multirow{3}{*}{SL + $W_1$} &1 & 63617 & 3810 & 2062 & 0.08 & 0.19 & \multirow{3}{*}{0.24}\\
			&& 2 & 11909 & 6480 & 9406  &0.77 & 0.55 & \\
			&& 3 & 2590 & 4047 & 35056 &0.16 & 0.25 &  \\\\ 
			& \multirow{3}{*}{SL0 + $W_2$} &1 & 49161&13433&6895 &0.29 & 0.24 & \multirow{3}{*}{0.39} \\ %
			&& 2 & 8453&9915&9427 &0.64 & 0.69 & \\
			&& 3 & 7104&8867&25723 &0.38 & 0.39 &  \\\\ 
			& \multirow{3}{*}{SL + $W_2$} &1 & 55798&12432&1259 & 0.20 & 0.11 & \multirow{3}{*}{0.26}\\ %
			&& 2 & 6166&14560&7069  &0.48 & 0.59 & \\
			&& 3 & 802&8541&32350 &0.22 & 0.20 &  \\ 
			\bottomrule
		\end{tabular}
        \end{threeparttable}
\end{center}
\end{table}
\FloatBarrier
\begin{table}[htb]
\footnotesize{}
\begin{center}
\caption*{Table S5. Simulation results for classification of the ordinal clinical significance of PCa assuming strong regional heterogeneity using QDA as the base learner.
\label{ordinal_simu_table_qda_strong}}
    \begin{threeparttable} 	
    \begin{tabular}{cccccccccc}
			\toprule
			\multirow{2}{*}{\makecell{Spatial\\Correlation}} & \multirow{2}{*}{Method} & \multirow{2}{*}{\makecell{True PCa\\Category}}
		    & \multicolumn{3}{c}{Classified Category} & \multirow{2}{*}{FPR} & \multirow{3}{*}{FDR} & \multirow{2}{*}{\makecell{Overall\\Error Rate}}\\
		    &&& 1 & 2 & 3 & & & \\
		    \midrule
			\multirow{20}{*}{\makecell{Moderate\\$\sigma^2=4$\\$\phi=0.2$\\$\nu=0.8$}}
			&\multirow{3}{*}{Baseline} &1 & 58424 & 1265 & 9287 & 0.15 & 0.38 & \multirow{3}{*}{0.41}\\ %
			&& 2 & 15978 & 1952 & 9660  &0.93 & 0.55 & \\
			&& 3 & 19307 & 1115 & 20964 &0.49 & 0.48 & \\\\ 
			& \multirow{3}{*}{SL0 + $W_1$} &1 & 60502&0&8075 & 0.12 & 0.34 & \multirow{3}{*}{0.37}\\ %
			&& 2 & 15865&0&11725 &1.00 & NA & \\
			&& 3 & 14704&0&26682 &0.36 & 0.43 &  \\\\ 
			&& 2 & 16347 & 0 & 11243  &1.00 & NA & \\
			&& 3 & 10320 & 0 & 31066 &0.25 & 0.37 &  \\\\ 
			& \multirow{3}{*}{SL0 + $W_2$} &1 & 48577&13453&6947 &0.30 & 0.24 & \multirow{3}{*}{0.40} \\ %
			&& 2 & 8433&9595&9562 &0.65 & 0.70 & \\
			&& 3 & 7236&9099&25051 &0.39 & 0.40 &  \\\\ 
			& \multirow{3}{*}{SL + $W_2$} &1 & 48046&16241&4690 & 0.30 & 0.20 & \multirow{3}{*}{0.37}\\ %
			&& 2 & 8343&10462&8785  &0.62 & 0.72 & \\
			&& 3 & 3449&10129&27808 &0.33 & 0.33 &  \\
			\midrule
			\multirow{20}{*}{\makecell{Strong\\$\sigma^2=10$\\$\phi=0.5$\\$\nu=1.5$}}
			& \multirow{3}{*}{Baseline} &1 & 58969 & 1544 & 9525 & 0.16 & 0.37 & \multirow{3}{*}{0.41}\\ %
			&& 2 & 16008 & 2342 & 9665  &0.92 & 0.56 & \\
			&& 3 & 19251 & 1374 & 21398 &0.49 & 0.47 & \\\\ 
			& \multirow{3}{*}{SL0 + $W_1$} &1 & 61263&0&8775 & 0.13 & 0.34 & \multirow{3}{*}{0.37}\\ %
			&& 2 & 16173&0&11842 &1.00 & NA & \\
			&& 3 & 15106&0&26917 &0.36 & 0.43 &  \\\\ 
			& \multirow{3}{*}{SL + $W_1$} &1 & 64151 & 3480&2407 & 0.08 & 0.20 & \multirow{3}{*}{0.25}\\ %
			&& 2 & 12569&5579&9868  &0.80 &0.57 & \\
			&& 3 & 3150&3789&35085 &0.16 & 0.26 &  \\\\ 
			& \multirow{3}{*}{SL0 + $W_2$} &1 &  49228&13479&7331 &0.30 & 0.25 & \multirow{3}{*}{0.40} \\ %
			&& 2 & 8584 & 9621 & 9809  &0.66 & 0.70 & \\
			&& 3 & 7515 & 8947 & 25561 &0.39 & 0.40 &  \\\\ 
			& \multirow{3}{*}{SL + $W_2$} &1 &  55376&13242&1420 & 0.21 & 0.12 & \multirow{3}{*}{0.27}\\ %
			&& 2 & 6405&14309&7301  &0.49 & 0.61 & \\
			&& 3 & 953&8972&32097 &0.24 & 0.21 &  \\ 
			\bottomrule
		\end{tabular}
        \end{threeparttable}
\end{center}
\end{table}
\FloatBarrier

\begin{table}[htb]
\footnotesize{}
\begin{center}
\caption*{Table S6. Simulation results for classification of the ordinal clinical significance of PCa assuming moderate regional heterogeneity using RF as the base learner.
\label{ordinal_simu_table_rf_mod}}
    \begin{threeparttable} 	
    \begin{tabular}{ccccccccc}
			\toprule
			\multirow{2}{*}{\makecell{Spatial\\Correlation}} & \multirow{2}{*}{Method} & \multirow{2}{*}{\makecell{True PCa\\Category}}
		    & \multicolumn{3}{c}{Classified Category} & \multirow{2}{*}{FPR} & \multirow{3}{*}{FDR} & \multirow{2}{*}{\makecell{Overall\\Error Rate}}\\
		    &&& 1 & 2 & 3 & & & \\
		    \midrule
			\multirow{20}{*}{\makecell{Moderate\\$\sigma^2=4$\\$\phi=0.2$\\$\nu=0.8$}}
			&\multirow{3}{*}{Baseline} &1 & 56143 & 4822 & 8679 & 0.19 & 0.33 & \multirow{3}{*}{0.39}\\ %
			&& 2 & 13031 & 6568 & 8258  &0.76 & 0.58 & \\
			&& 3 & 15148 & 4406 & 22234 &0.47 & 0.43 & \\\\ 
			& \multirow{3}{*}{SL0 + $W_1$} &1 & 60782 &0&8862 & 0.13 & 0.34 & \multirow{3}{*}{0.37}\\ %
			&& 2 & 16504 & 0 & 11354 &1.00 & NA & \\
			&& 3 & 15303 & 0 & 26484 &0.37 & 0.43 &  \\\\	
			& \multirow{3}{*}{SL + $W_1$} &1 & 62777 & 0 & 6867 & 0.10 & 0.30 & \multirow{3}{*}{0.32}\\ %
			&& 2 & 16573 & 0 & 11284  &1.00 & NA & \\
			&& 3 & 9982 & 0 & 31805 &0.24 & 0.36 &  \\\\ 
			& \multirow{3}{*}{SL0 + $W_2$} &1 & 48698&13242&7705 &0.30 & 0.26 & \multirow{3}{*}{0.40} \\ %
			&& 2 & 8884&9325&9649 &0.67 & 0.70 & \\
			&& 3 & 7962&8508&25317 &0.39 & 0.41 &  \\\\ 
			& \multirow{3}{*}{SL + $W_2$} &1 & 49647&15591&4406 & 0.29 & 0.19 & \multirow{3}{*}{0.36}\\ %
			&& 2 & 8367 & 10916 & 8574  &0.61 & 0.70 & \\
			&& 3 & 3225 & 10036 & 28226 &0.32 & 0.32 &  \\
			\midrule
			\multirow{20}{*}{\makecell{Strong\\$\sigma^2=10$\\$\phi=0.5$\\$\nu=1.5$}}
			&\multirow{3}{*}{Baseline} &1 & 55754 & 4903 & 8831 & 0.20 & 0.34 & \multirow{3}{*}{0.39}\\ %
			&& 2 & 12999 & 6626 & 8170  &0.76 & 0.59 & \\
			&& 3 & 15118 & 4443 & 22131 &0.47 & 0.43 & \\\\ 
			& \multirow{3}{*}{SL0 + $W_1$} &1 & 60582 &0&8907 & 0.13 & 0.34 & \multirow{3}{*}{0.37}\\
			&& 2 & 16499 & 0 & 11296 &1.00 & NA & \\
			&& 3 & 15317 & 0 & 26376 &0.37 & 0.43 &  \\\\	
			& \multirow{3}{*}{SL + $W_1$} &1 & 63664 & 3787 & 2038 & 0.08 & 0.18 & \multirow{3}{*}{0.24}\\ %
			&& 2 & 11869 & 6482 & 9444  &0.77 & 0.55 & \\
			&& 3 & 2550 & 4032 & 35111 &0.16 & 0.25 &  \\\\ 
			& \multirow{3}{*}{SL0 + $W_2$} &1 & 48635&13094&7760 &0.30 & 0.26 & \multirow{3}{*}{0.40} \\ %
			&& 2 & 8883&9230&9682 &0.67 & 0.70 & \\
			&& 3 & 7940&8335&25418 &0.39 & 0.41 &  \\\\ 
			& \multirow{3}{*}{SL + $W_2$} &1 & 56034&12213&1241 & 0.19 & 0.11 & \multirow{3}{*}{0.26}\\ %
			&& 2 & 6198&14511&7085  &0.48 & 0.59 & \\
			&& 3 & 798&8492&32403 &0.22 & 0.20 &  \\ 
			\bottomrule
		\end{tabular}
        \end{threeparttable}
\end{center}
\end{table}
\FloatBarrier
\begin{table}[htb]
\footnotesize{}
\begin{center}
\caption*{Table S7. Simulation results for classification of the ordinal clinical significance of PCa assuming strong regional heterogeneity using RF as the base learner.
\label{ordinal_simu_table_rf_strong}}
    \begin{threeparttable} 	
    \begin{tabular}{cccccccccc}
			\toprule
			\multirow{2}{*}{\makecell{Spatial\\Correlation}} & \multirow{2}{*}{Method} & \multirow{2}{*}{\makecell{True PCa\\Category}}
		    & \multicolumn{3}{c}{Classified Category} & \multirow{2}{*}{FPR} & \multirow{3}{*}{FDR} & \multirow{2}{*}{\makecell{Overall\\Error Rate}}\\
		    &&& 1 & 2 & 3 & & & \\
		    \midrule
			\multirow{20}{*}{\makecell{Moderate\\$\sigma^2=4$\\$\phi=0.2$\\$\nu=0.8$}}
			&\multirow{3}{*}{Baseline} &1 & 55253 & 4842 & 8882 & 0.20 & 0.34 & \multirow{3}{*}{0.39}\\ %
			&& 2 & 12821 & 6795 & 7973  &0.75 & 0.58 & \\
			&& 3 & 15366 & 4384 & 21635 &0.48 & 0.44 & \\\\ 
			& \multirow{3}{*}{SL0 + $W_1$} &1 & 60012&0&8965 & 0.13 & 0.35 & \multirow{3}{*}{0.38}\\ %
			&& 2 & 16254&0&11336 &1.00 & NA & \\
			&& 3 & 15401&0&25985 &0.37 & 0.44 &  \\\\ 
			& \multirow{3}{*}{SL + $W_1$} &1 & 61693 & 0 & 7284 & 0.11 & 0.30 & \multirow{3}{*}{0.33}\\ %
			&& 2 & 16267 & 0 & 11323  &1.00 & NA & \\
			&& 3 & 10211 & 0 & 31175 &0.25 & 0.37 &  \\\\ 
			& \multirow{3}{*}{SL0 + $W_2$} &1 & 48080&12942&7955 &0.30 & 0.26 & \multirow{3}{*}{0.41} \\ %
			&& 2 & 8862 & 8941 &9787 &0.68 & 0.70 & \\
			&& 3 & 8009&8413&24964 &0.40 & 0.42 &  \\\\ 
			& \multirow{3}{*}{SL + $W_2$} &1 & 47961&16150&4866 & 0.30 & 0.20 & \multirow{3}{*}{0.37}\\ %
			&& 2 & 8354&10350&8885  &0.62 & 0.72 & \\
			&& 3 & 3419&10004&27964 &0.32 & 0.33 &  \\ 
			\midrule
			\multirow{20}{*}{\makecell{Strong\\$\sigma^2=10$\\$\phi=0.5$\\$\nu=1.5$}}
			&\multirow{3}{*}{Baseline} &1 & 55815 & 5060 & 9164 & 0.20 & 0.34 & \multirow{3}{*}{0.40}\\ %
			&& 2 & 13061 & 6857 & 8097  &0.76 & 0.58 & \\
			&& 3 & 15544 & 4493 & 21986 &0.48 & 0.44 & \\\\ 
			& \multirow{3}{*}{SL0 + $W_1$} &1 & 60899 &0&9139 & 0.13 & 0.35 & \multirow{3}{*}{0.38}\\ %
			&& 2 & 16645&0&11370 &1.00 & NA & \\
			&& 3 & 15767&0&26256 &0.38 & 0.44 &  \\\\ 
			& \multirow{3}{*}{SL + $W_1$} &1 & 64120 & 3433&2485 & 0.08 & 0.19 & \multirow{3}{*}{0.25}\\ %
			&& 2 & 12469&5468&10078  &0.80 & 0.57 & \\
			&& 3 & 3028&3570&35425 &0.16 & 0.26 &  \\\\ 
			&\multirow{3}{*}{SL0 + $W_2$} &1 &  48754&13095&8189 &0.30 & 0.26 & \multirow{3}{*}{0.41} \\ %
			&& 2 & 9045 & 9046 & 9923  &0.68 & 0.70 & \\
			&& 3 & 8214 & 8339 & 25470 &0.39 & 0.42 &  \\\\ 
			& \multirow{3}{*}{SL + $W_2$} &1 &  55396&13190&1452 & 0.21 & 0.12 & \multirow{3}{*}{0.27}\\ %
			&& 2 & 6388&14191&7435  &0.49 & 0.61 & \\
			&& 3 & 933&8656&32435 &0.23 & 0.21 &  \\ 
			\bottomrule
		\end{tabular}
        \end{threeparttable}
\end{center}
\end{table}
\FloatBarrier

\begin{table}[htb]
\footnotesize{}
\begin{center}
\caption*{Table S8. Simulation results of GLM + QDA + RF for classifying the ordinal clinical significance of PCa assuming moderate regional heterogeneity.
\label{ordinal_simu_table_combine_mod}}
    \begin{threeparttable} 	
    \begin{tabular}{ccccccccc}
			\toprule
			\multirow{2}{*}{\makecell{Spatial\\Correlation}} & \multirow{2}{*}{Method} & \multirow{2}{*}{\makecell{True PCa\\Category}}
		    & \multicolumn{3}{c}{Classified Category} & \multirow{2}{*}{FPR} & \multirow{3}{*}{FDR} & \multirow{2}{*}{\makecell{Overall\\Error Rate}}\\
		    &&& 1 & 2 & 3 & & & \\
		    \midrule
			\multirow{16}{*}{\makecell{Moderate\\$\sigma^2=4$\\$\phi=0.2$\\$\nu=0.8$}}
		    & \multirow{3}{*}{SL0 + $W_1$} &1 & 61024&0&8621 & 0.12 & 0.33 & \multirow{3}{*}{0.36}\\ %
			&& 2 & 16180&0&11677 &1.00 & NA & \\
			&& 3 & 14085&0&27702 &0.34 & 0.42 &  \\\\ 
			& \multirow{3}{*}{SL + $W_1$} &1 & 63155 & 0 & 6489 & 0.09 & 0.29 & \multirow{3}{*}{0.31}\\ %
			&& 2 & 16440 & 0 & 11417  &1.00 & NA & \\
			&& 3 & 9206 & 0 & 32581 &0.22 & 0.35 &  \\\\ 
			& \multirow{3}{*}{SL0 + $W_2$} &1 & 48114&14220&7311 &0.31 & 0.23 & \multirow{3}{*}{0.39} \\ %
			&& 2 & 8102&10361&9395 &0.63 & 0.69 & \\
			&& 3 & 6584&8927&26276 &0.37 & 0.39 &  \\\\ 
			& \multirow{3}{*}{SL + $W_2$} &1 & 50473&15254&3918 & 0.28 & 0.18 & \multirow{3}{*}{0.35}\\ %
			&& 2 & 8164&11347&8347  &0.59 & 0.69 & \\
			&& 3 & 2890&10245&28651 &0.31 & 0.30 &  \\ 
			\midrule
			\multirow{16}{*}{\makecell{Strong\\$\sigma^2=10$\\$\phi=0.5$\\$\nu=1.5$}}
			& \multirow{3}{*}{SL0 + $W_1$} &1 & 60814&0&8675 & 0.12 & 0.33 & \multirow{3}{*}{0.36}\\ %
			&& 2 & 16169&0&11626 &1.00 & NA & \\
			&& 3 & 14171&0&27522 &0.34 & 0.42 &  \\\\ 
			& \multirow{3}{*}{SL + $W_1$} &1 & 63582 & 4570&1337 & 0.09 & 0.16 & \multirow{3}{*}{0.22}\\ %
			&& 2 & 10582&8792&8421  &0.68 & 0.51 & \\
			&& 3 & 1605&4696&35392 &0.15 & 0.22 &  \\\\ 
			& \multirow{3}{*}{SL0 + $W_2$} &1 & 48102&14000&7387 &0.31 & 0.23 & \multirow{3}{*}{0.39} \\ %
			&& 2 & 8122&10252&9421  &0.63 & 0.69 & \\
			&& 3 & 6620&8765&26308 &0.37 & 0.39 &  \\\\ 
			& \multirow{3}{*}{SL + $W_2$} &1 & 57296&11324&869 & 0.18 & 0.10 & \multirow{3}{*}{0.24}\\ %
			&& 2 & 5793&15441&6561  &0.44 & 0.55 & \\
			&& 3 & 552&7890&33251 &0.20 & 0.18 &  \\ 
			\bottomrule
		\end{tabular}
        \end{threeparttable}
\end{center}
\end{table}
\FloatBarrier

\begin{table}[htb]
\footnotesize{}
\begin{center}
\caption*{Table S9. Simulation results of GLM + QDA + RF for classifying the ordinal clinical significance of PCa assuming strong regional heterogeneity.
\label{ordinal_simu_table_combine_strong}}
    \begin{threeparttable} 	
    \begin{tabular}{ccccccccc}
			\toprule
			\multirow{2}{*}{\makecell{Spatial\\Correlation}} & \multirow{2}{*}{Method} & \multirow{2}{*}{\makecell{True PCa\\Category}}
		    & \multicolumn{3}{c}{Classified Category} & \multirow{2}{*}{FPR} & \multirow{3}{*}{FDR} & \multirow{2}{*}{\makecell{Overall\\Error Rate}}\\
		    &&& 1 & 2 & 3 & & & \\
		    \midrule
			\multirow{16}{*}{\makecell{Moderate\\$\sigma^2=4$\\$\phi=0.2$\\$\nu=0.8$}}
			& \multirow{3}{*}{SL0 + $W_1$} &1 & 60292&0&8685 & 0.13 & 0.33 & \multirow{3}{*}{0.37}\\ %
			&& 2 & 15868&0&11722 &1.00 & NA & \\
			&& 3 & 14320&0&27065 &0.35 & 0.43 &  \\\\ 
			& \multirow{3}{*}{SL + $W_1$} &1 & 62548 & 0 & 6429 & 0.09 & 0.29 & \multirow{3}{*}{0.31}\\ %
			&& 2 & 16196 & 1 & 11393  &1.00 & NA & \\
			&& 3 & 9104 & 0 & 32281 &0.22 & 0.36 &  \\\\ 
			& \multirow{3}{*}{SL0 + $W_2$} &1 & 47441&14005&7531 &0.31 & 0.24 & \multirow{3}{*}{0.40} \\ %
			&& 2 & 8083&9894&9613 &0.64 & 0.70 & \\
			&& 3 & 6673&8904&25809 &0.38 & 0.40 &  \\\\ 
			& \multirow{3}{*}{SL + $W_2$} &1 & 49966&15108&3903 & 0.28 & 0.18 & \multirow{3}{*}{0.35}\\ %
			&& 2 & 8093&11104&8393  &0.60 & 0.69 & \\
			&& 3 & 2888&10060&28438 &0.31 & 0.30 &  \\ 
			\midrule
			\multirow{16}{*}{\makecell{Strong\\$\sigma^2=10$\\$\phi=0.5$\\$\nu=1.5$}}
			& \multirow{3}{*}{SL0 + $W_1$} &1 & 61141&0&8897 & 0.13 & 0.34 & \multirow{3}{*}{0.37}\\ %
			&& 2 & 16245&0&11770 &1.00 & NA & \\
			&& 3 & 14734&0&27289 &0.35 & 0.43 &  \\\\ 
			& \multirow{3}{*}{SL + $W_1$} &1 & 64154 & 4545&1340 & 0.08 & 0.16 & \multirow{3}{*}{0.23}\\ %
			&& 2 & 10801&8734&8480  &0.69 & 0.51 & \\
			&& 3 & 1686&4679&35658 &0.15 & 0.22 &  \\\\ 
			& \multirow{3}{*}{SL0 + $W_2$} &1 & 48144&14103&7791 &0.31 & 0.24 & \multirow{3}{*}{0.40} \\ %
			&& 2 & 8256&10020&9738  &0.64 & 0.70 & \\
			&& 3 & 6937&8839&26247 &0.38 & 0.40 &  \\\\ 
			& \multirow{3}{*}{SL + $W_2$} &1 & 57678&11490&870 & 0.18 & 0.10 & \multirow{3}{*}{0.24}\\ %
			&& 2 & 5943&15471&6601  &0.45 & 0.56 & \\
			&& 3 & 586&7911&33525 &0.20 & 0.18 &  \\ 
			\bottomrule
		\end{tabular}
        \end{threeparttable}
\end{center}
\end{table}
\FloatBarrier

\subsection*{C.2 Application to Patient Data}
Tables S10 and S11 report the ordinal classification results on patient data using QDA and RF, respectively, as the base learner. 
Table S12 reports the ordinal classification results on patient data when combining the multi-resolution GLM, QDA and RF together. 
Each Table reports the results using either the predicted probabilities for the first two categories or the classified cancer categories from the stage-one multi-resolution base learners as the covariates for the stage-two model in the proposed super learner.
\begin{table}[htb]
\footnotesize{}
\begin{center}
\caption*{Table S10. Ordinal classification results on patient data using QDA as the base learner.
\label{ordinal_application_table_qda}
}
    \begin{threeparttable} 	
    \begin{tabular}{cccccccccc}
			\toprule
			\multirow{2}{*}{\makecell{Stage-one\\Output}} & \multirow{2}{*}{Method} & \multirow{2}{*}{\makecell{True PCa\\Category}}
		    & \multicolumn{3}{c}{Classified Category} & \multirow{2}{*}{FPR} & \multirow{3}{*}{FDR} & \multirow{2}{*}{\makecell{Overall\\Error Rate}}\\
		    &&& 1 & 2 & 3 & & & \\
		    \midrule
			\multirow{21}{*}{\makecell{Probabilities\\for the First\\Two Categories}}
			&&& 1 & 2 & 3 &&& \\		    
			& \multirow{3}{*}{Baseline} &1 & 87136 &297&1554 & 0.021 & 0.152 & \multirow{3}{*}{0.168}\\ 
			&& 2 & 5891&10&342  &0.998 & 0.975 & \\
			&& 3 & 9782&90&1835 &0.843 & 0.508 &  \\\\ 
			& \multirow{3}{*}{SL0 + $W_1$} &1 & 87515 &0&1472 & 0.017 & 0.155 & \multirow{3}{*}{0.167}\\ %
			&& 2 & 5937&0&306  &1.000 & NA & \\
			&& 3 & 10136&0&1571 &0.866 & 0.531 &  \\\\ 
			& \multirow{3}{*}{SL + $W_1$} &1 & 87952&0&1035 & 0.012 & 0.146 & \multirow{3}{*}{0.152}\\ %
			&& 2 & 5984&0&259  &1.000 & NA & \\
			&& 3 & 9011&0&2696 &0.770 & 0.324 &  \\\\ 
			& \multirow{3}{*}{SL0 + $W_2$} &1 & 65662&13795&9530 & 0.262 & 0.086 & \multirow{3}{*}{0.321}\\ %
			&& 2 & 2484&1312&2447  &0.790 & 0.925 & \\
			&& 3 & 3702&2344&5661 &0.516 & 0.679 &  \\\\ 
			& \multirow{3}{*}{SL + $W_2$} &1 & 64615&17302&7070 &0.274 & 0.080 & \multirow{3}{*}{0.324} \\ %
			&& 2 & 2590&1733&1920  &0.722 & 0.921 & \\
			&& 3 & 3000&2768&5939 &0.493 & 0.602 &  \\ 
			\midrule
			\multirow{21}{*}{\makecell{Categories}}
			& \multirow{3}{*}{Baseline} &1 & 87136 &297&1554 & 0.021 & 0.152 & \multirow{3}{*}{0.168}\\ 
			&& 2 & 5891&10&342  &0.998 & 0.975 & \\
			&& 3 & 9782&90&1835 &0.843 & 0.508 &  \\\\ 
			& \multirow{3}{*}{SL0 + $W_1$} &1 & 87431 &0&1556 & 0.017 & 0.150 & \multirow{3}{*}{0.163}\\ %
			&& 2 & 5846&0&397  &1.000 & NA & \\
			&& 3 & 9582&0&2125 &0.818 & 0.479 &  \\\\ 
			& \multirow{3}{*}{SL + $W_1$} &1 & 87684&0&1303 & 0.015 & 0.146 & \multirow{3}{*}{0.153}\\ %
			&& 2 & 6111&0&132  &1.000 & NA & \\
			&& 3 & 8842&0&2865 &0.755 & 0.334 &  \\\\ 
			& \multirow{3}{*}{SL0 + $W_2$} &1 & 82439&1400&5148 & 0.074 & 0.118 & \multirow{3}{*}{0.182}\\ %
			&& 2 & 4345&527&1371  &0.916 & 0.779 & \\
			&& 3 & 6696&460&4551 &0.611 & 0.589 &  \\\\ 
			& \multirow{3}{*}{SL + $W_2$} &1 & 76014&8874&4099 &0.146 & 0.088 & \multirow{3}{*}{0.227} \\ %
			&& 2 & 3397&1764&1082  &0.717 & 0.870 & \\
			&& 3 & 3915&2946&4846 &0.586 & 0.517 &  \\ 
			\bottomrule
		\end{tabular}
        \end{threeparttable}
\end{center}
\end{table}
\FloatBarrier

\begin{table}[htb]
\footnotesize{}
\begin{center}
\caption*{Table S11. Ordinal classification results on patient data using RF as the base learner.
\label{ordinal_application_table_rf}
}
    \begin{threeparttable} 	
    \begin{tabular}{cccccccccc}
			\toprule
			\multirow{2}{*}{\makecell{Stage-one\\Output}} & \multirow{2}{*}{Method} & \multirow{2}{*}{\makecell{True PCa\\Category}}
		    & \multicolumn{3}{c}{Classified Category} & \multirow{2}{*}{FPR} & \multirow{3}{*}{FDR} & \multirow{2}{*}{\makecell{Overall\\Error Rate}}\\
		    &&& 1 & 2 & 3 & & & \\
		    \midrule
			\multirow{21}{*}{\makecell{Probabilities\\for the First\\Two Categories}}
			& \multirow{3}{*}{Baseline} &1 & 84951 &1021&3015 & 0.045 & 0.143 & \multirow{3}{*}{0.180}\\ %
			&& 2 & 5511&169&563  &0.973 & 0.898 & \\
			&& 3 & 8719&470&2518 &0.785 & 0.587 &  \\\\ 
			& \multirow{3}{*}{SL0 + $W_1$} &1 & 87393 &0&1594 & 0.018 & 0.146 & \multirow{3}{*}{0.158}\\ %
			&& 2 & 5833&0&410  &1.000 & NA & \\
			&& 3 & 9113&0&2594 &0.778 & 0.436 &  \\\\ 
			& \multirow{3}{*}{SL + $W_1$} &1 & 87309&0&1678 & 0.019 & 0.125 & \multirow{3}{*}{0.134}\\ %
			&& 2 & 5970&0&273  &1.000 & NA & \\
			&& 3 & 6456&0&5251 &0.551 & 0.271 &  \\\\ 
			& \multirow{3}{*}{SL0 + $W_2$} &1 & 65908&13174&9905 & 0.259 & 0.086 & \multirow{3}{*}{0.311}\\ %
			&& 2 & 3206&1291&1746  &0.793 & 0.923 & \\
			&& 3 & 2977&2385&6345 &0.458 & 0.647 &  \\\\ 
			& \multirow{3}{*}{SL + $W_2$} &1 & 64888&17586&6513 &0.271 & 0.050 & \multirow{3}{*}{0.295} \\ %
			&& 2 & 2304&2545&1394  &0.592 & 0.888 & \\
			&& 3 & 1100&2626&7981 &0.318 & 0.498 &  \\ 
			\midrule
			\multirow{21}{*}{Categories}
			& \multirow{3}{*}{Baseline} &1 & 84927 &1037&3023 & 0.046 & 0.143 & \multirow{3}{*}{0.180}\\ %
			&& 2 & 5475&172&596  &0.972 & 0.898 & \\
			&& 3 & 8671&481&2555 &0.782 & 0.586 &  \\\\ 
			& \multirow{3}{*}{SL0 + $W_1$} &1 & 87495 &0&1492 & 0.017 & 0.149 & \multirow{3}{*}{0.160}\\ %
			&& 2 & 5964&0&279  &1.000 & NA & \\
			&& 3 & 9363&0&2344 &0.800 & 0.430 &  \\\\ 
			& \multirow{3}{*}{SL + $W_1$} &1 & 87515&0&1472 & 0.017 & 0.132 & \multirow{3}{*}{0.139}\\ %
			&& 2 & 6115&0&128  &1.000 & NA & \\
			&& 3 & 7197&0&4510 &0.615 & 0.262 &  \\\\ 
			& \multirow{3}{*}{SL0 + $W_2$} &1 & 78988&2218&7781 & 0.112 & 0.115 & \multirow{3}{*}{0.209}\\ %
			&& 2 & 4558&305&1380  &0.951 & 0.903 & \\
			&& 3 & 5683&687&5337 &0.544 & 0.633 &  \\\\ 
			& \multirow{3}{*}{SL + $W_2$} &1 & 70088&13920&4979 &0.212 & 0.064 & \multirow{3}{*}{0.263} \\ %
			&& 2 & 3199&1770&1274  &0.716 & 0.906 & \\
			&& 3 & 1610&3141&6956 &0.406 & 0.473 &  \\ 
			\bottomrule
		\end{tabular}
        \end{threeparttable}
\end{center}
\end{table}
\FloatBarrier

\begin{table}[htb]
\footnotesize{}
\begin{center}
\caption*{Table S12. Ordinal classification results for GLM + QDA + RF on patient data. 
\label{ordinal_application_table_combine}
}
    \begin{threeparttable} 	
    \begin{tabular}{cccccccccc}
			\toprule
			\multirow{2}{*}{\makecell{Stage-one\\Output}} & \multirow{2}{*}{Method} & \multirow{2}{*}{\makecell{True PCa\\Category}}
		    & \multicolumn{3}{c}{Classified Category} & \multirow{2}{*}{FPR} & \multirow{3}{*}{FDR} & \multirow{2}{*}{\makecell{Overall\\Error Rate}}\\
		    &&& 1 & 2 & 3 & & & \\
		    \midrule
			\multirow{17}{*}{\makecell{Probabilities\\for the First\\Two Categories}}
			& \multirow{3}{*}{SL0 + $W_1$} &1 & 86704 &0&2283 & 0.026 & 0.141 & \multirow{3}{*}{0.158}\\ %
			&& 2 & 5865&0&378  &1.000 & NA & \\
			&& 3 & 8381&0&3326 &0.716 & 0.444 &  \\\\ 
			& \multirow{3}{*}{SL + $W_1$} &1 & 86358&0&2629 & 0.030 & 0.125 & \multirow{3}{*}{0.143}\\ %
			&& 2 & 6023&0&220  &1.000 & NA & \\
			&& 3 & 6369&0&5338 &0.544 & 0.348 &  \\\\ 
			& \multirow{3}{*}{SL0 + $W_2$} &1 & 66003&15079&7905 & 0.258 & 0.076 & \multirow{3}{*}{0.304}\\ %
			&& 2 & 2487&2004&1752  &0.679 & 0.897 & \\
			&& 3 & 2931&2318&6458 &0.448 & 0.599 &  \\\\ 
			& \multirow{3}{*}{SL + $W_2$} &1 & 66130&16911&5946 &0.257 & 0.058 & \multirow{3}{*}{0.290} \\ %
			&& 2 & 2600&2826&817 &0.547 & 0.877 & \\
			&& 3 & 1456&3266&6985 &0.403 & 0.492 &  \\ 
			\midrule
			\multirow{17}{*}{Categories}
			& \multirow{3}{*}{SL0 + $W_1$} &1 & 87613 &0&1374 & 0.015 & 0.146 & \multirow{3}{*}{0.156}\\ %
			&& 2 & 5944&0&299  &1.000 & NA & \\
			&& 3 & 9016&0&2691 &0.770 & 0.383 &  \\\\ 
			& \multirow{3}{*}{SL + $W_1$} &1 & 86699&0&2288 & 0.026 & 0.131 & \multirow{3}{*}{0.145}\\ %
			&& 2 & 6090&0&153  &1.000 & NA & \\
			&& 3 & 6989&0&4718 &0.597 & 0.341 &  \\\\ 
			& \multirow{3}{*}{SL0 + $W_2$} &1 & 78898&4093&5996 & 0.113 & 0.109 & \multirow{3}{*}{0.206}\\ %
			&& 2 & 4395&770&1078  &0.877 & 0.873 & \\
			&& 3 & 5263&1192&5252 &0.551 & 0.574 &  \\\\ 
			& \multirow{3}{*}{SL + $W_2$} &1 & 71464&13560&3963 &0.219 & 0.077 & \multirow{3}{*}{0.266} \\ %
			&& 2 & 3573&1909&761  &0.662 & 0.893 & \\
			&& 3 & 2664&2589&6454 &0.411 & 0.418 &  \\ 
			\bottomrule
		\end{tabular}
        \end{threeparttable}
\end{center}
\end{table}
\FloatBarrier

\label{lastpage}